\documentclass{article}

\usepackage{microtype}
\usepackage{graphicx}
\usepackage{subcaption}
\usepackage{booktabs} 
\graphicspath{{figures/}{figures/LRA/}{figures/multiscale/}}

\usepackage{hyperref}

\usepackage{float}
\usepackage{tikz}
\usepackage{multirow}
\usepackage{tabularx}
\usepackage{array}

\usepackage[preprint]{icml2026}

\usepackage{amsmath}
\usepackage{amssymb}
\usepackage{mathtools}
\usepackage{amsthm}

\usepackage[capitalize,noabbrev]{cleveref}

\theoremstyle{plain}
\newtheorem{theorem}{Theorem}[section]
\newtheorem{proposition}[theorem]{Proposition}
\newtheorem{lemma}[theorem]{Lemma}
\newtheorem{corollary}[theorem]{Corollary}
\theoremstyle{definition}
\newtheorem{definition}[theorem]{Definition}
\newtheorem{assumption}[theorem]{Assumption}
\theoremstyle{remark}

\usepackage[disable,textsize=tiny]{todonotes}

\icmltitlerunning{Where to Add PDE Diffusion in Transformers}

\begin{document}

\twocolumn[
  \icmltitle{Where to Add PDE Diffusion in Transformers}




\icmlsetsymbol{equal}{*}

\begin{icmlauthorlist}
  \icmlauthor{Yukun Zhang}{equal,cuhk}
  \icmlauthor{Xueqing Zhou}{equal,fudan}
\end{icmlauthorlist}

\icmlaffiliation{cuhk}{The Chinese University of Hong Kong, Hong Kong, China}
\icmlaffiliation{fudan}{Fudan University, Shanghai, China}

\icmlcorrespondingauthor{Yukun Zhang}{215010026@link.cuhk.edu.cn}
\icmlcorrespondingauthor{Xueqing Zhou}{19210240101@fudan.edu.cn}

  \icmlkeywords{Transformers, locality, PDE diffusion, operator theory, spectral stability}

  \vskip 0.3in
]



\printAffiliationsAndNotice{}  

\begin{abstract}
Transformers achieve powerful content-based global routing via self-attention, but they do not explicitly encode a local geometric prior along the sequence axis, making the placement of locality-inducing modules largely empirical in prior hybrid designs. We study a simple, deterministic \emph{PDE diffusion} layer—one explicit Euler step of 1D heat-equation smoothing implemented by a discrete Neumann Laplacian under a spectral stability budget—and ask a structural question: \emph{where should diffusion be inserted relative to attention?} Our central claim is that diffusion and attention generally \emph{do not commute}, so inserting the same local operator before versus after attention can induce qualitatively different behaviors. To formalize this, we develop a three-layer operator-theoretic framework that (i) provides unconditional guarantees for the diffusion subsystem (spectral non-expansiveness and monotone Dirichlet-energy dissipation under $\alpha<1/2$), (ii) derives compositional perturbation bounds linking insertion effects to representation roughness and downstream amplification, and (iii) uses diffusion--attention non-commutativity as a diagnostic for structural “double-mixing” conflicts. Guided by these predictions, we evaluate seven canonical insertion positions on Long Range Arena and find that early diffusion acts as effective pre-regularization, improving average accuracy by $+4.1$ points when applied after embedding, whereas post-attention diffusion degrades performance by $-2.5$ points, consistent with the predicted conflict. We further show that a multi-scale diffusion variant yields consistent gains under the same global stability budget. Beyond this specific module, our analysis provides a transferable template for reasoning about local--global compositions in sequence models by cleanly separating provable guarantees, compositional bounds, and mechanistic diagnostics.
\end{abstract}

\section{Introduction}
\label{sec:intro}

\subsection{Motivation}

Transformers have become the dominant architecture for sequence modeling, largely due to self-attention's ability to perform content-based global routing~\citep{vaswani2017attention}. By enabling interactions between arbitrary token pairs, attention efficiently captures long-range dependencies and global semantic structure. However, standard Transformers lack an explicit \emph{local geometric prior} along the sequence dimension: they do not directly encourage neighboring token representations to vary smoothly or preserve local coherence. While attention can learn locality implicitly, it is not encoded architecturally and may be brittle under distribution shifts (e.g., different sequence lengths).

This gap has motivated hybrid designs that inject locality through convolutional layers~\citep{wu2019pay,dai2020funnel}, sparse patterns~\citep{beltagy2020longformer}, or other structured inductive biases. Yet a surprisingly basic question remains under-theorized: \emph{where should a local operator be placed relative to attention?} In practice, placement is often treated as a hyperparameter and tuned by validation performance. In this work, we ask whether placement can be predicted from structure rather than guessed from sweeps: \textbf{can operator theory explain when local smoothing helps and when it hurts?}

Our central observation is simple but consequential: local smoothing and content-based global routing are, in general, \emph{non-commuting operations}. Let $\mathcal{S}$ denote a local PDE diffusion (smoothing) operator and $\mathcal{A}$ an attention-based routing operator. Typically, $\mathcal{S} \circ \mathcal{A} \neq \mathcal{A} \circ \mathcal{S}$, and the commutator $[\mathcal{S}, \mathcal{A}] = \mathcal{S}\mathcal{A} - \mathcal{A}\mathcal{S}$ captures this incompatibility. This has direct architectural implications: applying PDE diffusion before attention can pre-regularize a rough representation space and provide a smoother manifold for routing; applying PDE diffusion after attention can instead blur structures just created by non-local mixing, inducing a harmful \emph{double-mixing} effect.

\subsection{Our Approach: A Three-Layer Operator-Theoretic Framework}

To formalize these intuitions, we develop a systematic operator-theoretic framework for analyzing interactions between locality and attention in Transformers. The framework is organized into three conceptual layers, each addressing a different aspect of the problem.

\paragraph{Layer I: Intrinsic stability (provable guarantees).}
We first study the PDE diffusion operator in isolation. Using a discrete Laplacian with a strict spectral budget constraint ($\alpha < 1/2$), we establish unconditional guarantees for the PDE diffusion subsystem alone. Specifically, we show that the operator is spectrally non-expansive, with strict contraction on non-constant modes, and that it induces monotone dissipation of the discrete Dirichlet energy. These results formalize PDE diffusion as a stable geometric low-pass filter that reduces local roughness independently of the Transformer's non-convex components, thereby providing a mathematically grounded stability anchor.
\emph{Takeaway: under a simple spectral budget, PDE diffusion is a certified non-expansive local smoother.}

\paragraph{Layer II: Compositional analysis (perturbation bounds).}
We then analyze how this stable PDE diffusion operator interacts with the rest of the Transformer when composed at different locations. Treating downstream modules—such as LayerNorm, attention, and MLPs—as locally Lipschitz mappings, we derive perturbation propagation bounds showing that the effect of inserting PDE diffusion at a given position depends on two factors: the geometric roughness of the representation at the insertion point, and the sensitivity of subsequent layers to perturbations. This analysis explains why the same PDE diffusion operator can be beneficial when applied early, where representations are rough and regularization is helpful, yet detrimental when applied late, where representations are already smooth and additional smoothing becomes unnecessary or harmful.
\emph{Takeaway: PDE diffusion helps most where representations are rough and downstream amplification is limited---typically early in the network.}

\paragraph{Layer III: Non-commutativity diagnostics (design guidelines).}
Finally, we turn to the interaction between PDE diffusion and attention. We formalize their incompatibility through the commutator $[\mathcal{S}, \mathcal{A}](X)$, which measures the representational discrepancy between pre-attention and post-attention diffusion. Under a local linearization of attention, this conflict scales with the mismatch between the local geometric structure enforced by PDE diffusion and the non-local routing patterns produced by attention. This perspective yields concrete, testable design guidelines: PDE diffusion should be applied early as a form of semantic manifold pre-regularization; post-attention PDE diffusion should be avoided due to structural conflict; and multi-scale PDE diffusion, when constrained by a global spectral budget, can improve robustness by covering multiple frequency bands without sacrificing stability.
\emph{Takeaway: because PDE diffusion and attention generally do not commute, post-attention smoothing can erase structures created by global routing.}

\subsection{Contributions}

This paper makes four main contributions. First, we develop a three-layer operator-theoretic framework that separates provable intrinsic guarantees from compositional perturbation analysis and architectural design heuristics, clarifying what can and cannot be established rigorously. Second, we derive principled architectural guidelines for inserting locality-inducing modules into Transformers, predicting that early-stage insertion should systematically outperform post-attention insertion due to fundamental operator incompatibility. Third, we provide controlled empirical validation on the Long Range Arena benchmark~\citep{tay2020long}, systematically evaluating seven insertion positions. The results align closely with theory: PDE diffusion after embedding yields a $+4.1$ percentage point improvement, while PDE diffusion after attention leads to a $-2.5$ point degradation. Finally, beyond the specific PDE diffusion operator studied here, we demonstrate how operator-theoretic reasoning can guide the design of hybrid architectures, offering a transferable methodology applicable to other local–global composition problems, such as convolution–attention or state-space–attention hybrids.

\subsection{Roadmap}

The remainder of the paper is organized as follows. Section~\ref{sec:related_work} reviews related hybrid architectures and theoretical perspectives. Section~\ref{sec:framework} introduces the operator-theoretic framework and defines the PDE diffusion layer precisely. Section~\ref{sec:theory} presents the three-layer theoretical analysis, including intrinsic stability guarantees and compositional perturbation bounds. Section~\ref{sec:experiments} empirically validates the resulting architectural guidelines on Long Range Arena. Section~\ref{sec:conclusion} concludes.

\section{Related Work}
\label{sec:related_work}

Our work sits at the intersection of (i) long-sequence Transformers, (ii) locality-inducing inductive biases, and (iii) stability- and operator-motivated architectural analysis. We position our contribution along three axes: we \emph{do not} change attention's routing rule or complexity (unlike efficient-attention variants), we use PDE diffusion as a \emph{fixed-form, spectrally-budgeted} local prior (not to be confused with generative diffusion models), and we provide a \emph{structural} explanation for insertion-position sensitivity (often treated empirically in hybrid designs).

\subsection{Long-Sequence Transformers}

The quadratic complexity of self-attention has motivated extensive work on efficient long-sequence Transformers. Representative directions include sparse patterns (e.g., Longformer~\citep{beltagy2020longformer}, BigBird~\citep{zaheer2020bigbird}, Sparse Transformers~\citep{child2019sparse}, Reformer~\citep{kitaev2020reformer}), low-rank or kernelized approximations (e.g., Performer~\citep{choromanski2021rethinking}, Linear Transformers~\citep{katharopoulos2020transformers}, Linformer~\citep{wang2020linformer}, Nystr\"omformer~\citep{xiong2021nystromformer}), and memory-optimized exact attention (e.g., FlashAttention~\citep{dao2022flashattention}). LRA~\citep{tay2020long,tay2021longrange} further highlights that efficiency improvements do not automatically translate into robust long-context generalization.

Our work is orthogonal: we keep attention intact and instead study how to inject an explicit \emph{local geometric prior} via a lightweight PDE diffusion operator. Because the PDE diffusion layer is linear-time, it can be combined with sparse/linear attention mechanisms as a complementary design component.

\subsection{Diffusion, PDEs, and Neural Operators}

PDE- and diffusion-inspired views have a long history in deep learning, including stability-aware architectures via dynamical-systems discretizations~\citep{haber2017stable,ruthotto2017deep} and classical diffusion for denoising/smoothing~\citep{weickert1998anisotropic}. More recently, neural operator learning aims to approximate families of PDE solution operators~\citep{li2021fourier}, and generative diffusion models treat diffusion as a stochastic generative process~\citep{ho2020denoising}.

Our use of PDE diffusion differs in both goal and modeling choice. We use a \emph{fixed-form} discrete Laplacian with a strict spectral budget to inject a local smoothness prior with unconditional stability guarantees, rather than learning an operator approximator or modeling a generative diffusion process. The resulting module is a deterministic architectural regularizer that targets the sequence dimension and integrates naturally with Transformers.

\subsection{Spectral Stability and Regularization in Deep Networks}

Stability and spectral control have been recognized as central issues in deep learning, particularly for very deep or long-context models.
Spectral normalization~\citep{miyato2018spectral} and Lipschitz-constrained architectures aim to control gradient explosion and improve robustness.
Layer normalization~\citep{ba2021layer} and residual connections~\citep{he2016deep} provide fundamental stability mechanisms in deep networks.
In Transformers, prior work has analyzed the instability of attention maps and the amplification of high-frequency components across layers.

Our work contributes to this line by introducing a simple yet effective mechanism for spectral control along the \emph{sequence dimension}.
By leveraging the discrete Laplacian with a strict spectral budget, we guarantee non-expansiveness and monotone Dirichlet energy dissipation for the diffusion subsystem.
Importantly, our analysis does not rely on global Lipschitz assumptions for the full Transformer; instead, it isolates what can be proven unconditionally and studies interactions with attention through controlled perturbation and non-commutativity.

\subsection{Position Sensitivity and Hybrid Architectures}

Hybrid architectures that combine local and global operations are increasingly common, including convolution--attention hybrids~\citep{wu2019pay,dai2020funnel} and alternatives to attention such as structured state space models~\citep{gu2022s4,gu2023mamba}. In many such designs, the placement of local operators relative to global mixing is decided empirically.

Our contribution is a principled account of this placement sensitivity. By framing diffusion and attention as generally non-commuting operators, we explain why \emph{pre-routing} diffusion can act as beneficial manifold regularization while \emph{post-routing} diffusion can introduce a destructive double-mixing effect. This operator-theoretic lens is not specific to diffusion and can serve as a transferable template for reasoning about other local--global compositions.

\section{Operator Framework and Diffusion Layer Design}
\label{sec:framework}

\subsection{Scope of Theoretical Guarantees}
\label{sec:guarantees}

This section introduces an operator-theoretic framework centered on a PDE diffusion layer and clarifies the precise scope of its theoretical guarantees. Our analysis is intentionally structured into three conceptual layers, each corresponding to a different level of rigor and assumption.

At the first level, we establish \emph{intrinsic stability properties} of the diffusion operator itself. These include spectral non-expansiveness and monotone energy dissipation, which hold unconditionally and independently of all other Transformer components. These results are provable without assumptions on attention, normalization, or nonlinear activations.

At the second level, we study how diffusion interacts compositionally with backbone modules such as LayerNorm, self-attention, and MLP blocks. This analysis relies on local Lipschitz assumptions and yields perturbation propagation bounds that explain why the effect of diffusion depends strongly on its insertion position within the network.

At the third level, we provide a \emph{diagnostic and heuristic analysis} based on the non-commutativity between diffusion and attention operators. This layer depends on a local linearization of attention and does not claim strict guarantees. Instead, it offers principled architectural guidance by quantifying geometric conflict between local smoothing and global routing.

This layered structure explicitly separates what is rigorously provable from what requires additional assumptions, thereby avoiding overstatement of theoretical claims.

\subsection{Core Operator Definitions}
\label{sec:operators}

We consider a sequence of length $L$ with hidden dimension $d$. Let $X \in \mathbb{R}^{L \times d}$ denote a layer representation, where $X_i \in \mathbb{R}^d$ is the feature vector at position $i$.

\paragraph{Boundary conditions.}
To preserve information at sequence boundaries, we impose Neumann (zero-flux) boundary conditions via endpoint replication,
\[
x_0 = x_1, \qquad x_{L+1} = x_L .
\]

\paragraph{Discrete Neumann Laplacian.}
For a vector $x \in \mathbb{R}^L$, the discrete Neumann Laplacian $\Delta_N$ acts elementwise as
\begin{equation}
(\Delta_N x)_i =
\begin{cases}
x_2 - x_1, & i = 1, \\
x_{i-1} - 2x_i + x_{i+1}, & 2 \le i \le L-1, \\
x_{L-1} - x_L, & i = L .
\end{cases}
\label{eq:neumann_laplacian}
\end{equation}
For matrix inputs, the operator is applied channel-wise along the sequence dimension, i.e., $(\Delta_N X)_{\cdot,j} = \Delta_N(X_{\cdot,j})$.

\paragraph{Single-scale diffusion.}
The minimal diffusion module is defined as
\[
\mathcal{S}_\alpha(X) = (I + \alpha \Delta_N) X,
\qquad 0 \le \alpha < \tfrac{1}{2}.
\]
The constraint $\alpha < \tfrac{1}{2}$ acts as a \emph{stability budget}, analogous to the CFL condition in explicit diffusion schemes, and guarantees spectral non-expansiveness and energy dissipation.

\paragraph{Multi-scale diffusion.}
To accommodate heterogeneous feature lengths, we extend the operator to a multi-scale formulation by combining $K$ strided Laplacians with step sizes $h_k$:
\[
\mathcal{S}_{\boldsymbol{\alpha}}(X)
= X + \sum_{k=1}^K \alpha_k \Delta_{N,h_k} X .
\]
A global budget constraint,
\[
\sum_{k=1}^K \alpha_k < \tfrac{1}{2},
\]
ensures that the same stability guarantees are preserved.

\subsection{Systematic Insertion Design Space}
\label{sec:positions}

We define a systematic design space consisting of seven canonical insertion strategies, ranging from early-stage pre-regularization to late-stage post-routing smoothing. These positions are summarized in Table~\ref{tab:positions}.

\begin{table*}[t]
\centering
\small
\setlength{\tabcolsep}{6pt}
\caption{Canonical insertion strategies for PDE diffusion layers, spanning the full two-column width.}
\label{tab:positions}
\begin{tabular}{c l l l}
\toprule
ID & Strategy & Composition & Theoretical Expectation \\
\midrule
\textbf{P1} & After-Embedding &
$X \leftarrow \mathcal{S}(\mathrm{Embed}(\cdot))$ &
\textbf{Optimal: semantic manifold pre-regularization} \\

P2 & After-MLP &
$X \leftarrow \mathcal{S}(X + \mathrm{MLP}(X))$ &
Moderate: partial regularization \\

P3 & Layer-Diffusion &
$X_{\ell+1} \leftarrow \mathcal{S}(\mathrm{Block}(X_\ell))$ &
Moderate: inter-layer smoothing \\

P4 & Before-LN &
$X \leftarrow \mathrm{LN}(\mathcal{S}(X))$ &
Moderate: interaction with normalization \\

P5 & In-Attention &
$\mathrm{Attn}(Q,K,V)\!\rightarrow\!\mathrm{Attn}(Q,K,\mathcal{S}(V))$ &
Unclear: internal intervention \\

P6 & Head-Diffusion &
$H \leftarrow \mathcal{S}_{\text{head}}(H)$ &
Unclear: cross-head exchange \\

\textbf{P7} & After-Attention &
$X \leftarrow \mathcal{S}(X + \mathrm{Attn}(X))$ &
\textbf{Worst: double-mixing conflict} \\
\bottomrule
\end{tabular}
\end{table*}

The central contrast lies between early and late insertion. Strategy P1 smooths representations before global routing, allowing attention to aggregate information on a less noisy semantic manifold. In contrast, P7 applies local smoothing after non-local mixing, potentially blurring structures that attention has just constructed.

\subsection{Implementation Details}
\label{sec:implementation}

The PDE diffusion operator has linear complexity in both time and memory. A single-scale diffusion requires $\mathcal{O}(Ld)$ operations, while the multi-scale variant scales as $\mathcal{O}(KLd)$.

To enforce the global stability budget exactly during training without manual clipping, we reparameterize the diffusion weights as
\[
\alpha_k = \tfrac{1}{2} \, \sigma(\eta) \, \mathrm{softmax}(\boldsymbol{\theta})_k,
\]
where $\eta \in \mathbb{R}$ and $\boldsymbol{\theta} \in \mathbb{R}^K$ are unconstrained learnable parameters.

Depending on architectural choice, diffusion can be applied either directly or followed by normalization,
\[
\text{Out} = \mathcal{S}(X),
\qquad \text{or} \qquad
\text{Out} = \mathrm{LN}(\mathcal{S}(X)).
\]

\subsection{Analysis Quantities: Roughness and Conflict Diagnostics}
\label{sec:diagnostics}

\paragraph{Local roughness.}
We quantify local geometric irregularity along the sequence dimension using the discrete Dirichlet energy,
\[
\mathcal{E}(X)
= \tfrac{1}{2} \langle X, -\Delta_N X \rangle
= \tfrac{1}{2} \sum_{j=1}^d \sum_{i=1}^{L-1}
\bigl(X_{i+1,j} - X_{i,j}\bigr)^2 .
\]
This quantity measures how rapidly representations vary between neighboring tokens.

\paragraph{Geometric conflict.}
To diagnose incompatibility between diffusion and attention, we define the diffusion--attention commutator,
\[
[\mathcal{S}, \mathcal{A}](X)
= \mathcal{S}(\mathcal{A}(X)) - \mathcal{A}(\mathcal{S}(X)).
\]
This operator measures the representational discrepancy between pre-attention and post-attention diffusion. In this work, we use it as a conceptual diagnostic for architectural compatibility (Layer~III), rather than as a measured training-time signal.

\section{Operator-Theoretic Analysis}
\label{sec:theory}

This section develops an operator-theoretic account of how locality-inducing PDE diffusion interacts with self-attention. To avoid overclaiming, we separate results into three layers: (i) unconditional properties of the PDE diffusion operator in isolation; (ii) compositional perturbation bounds under standard local regularity assumptions for Transformer modules; and (iii) a diagnostic interpretation of diffusion--attention non-commutativity based on a local linearization of attention. Full proofs are deferred to Appendix~\ref{app:proofs}.

\subsection{Layer I: Intrinsic Stability of Diffusion}
\label{subsec:intrinsic_stability}

We begin with properties that hold independently of the Transformer backbone. Let $\Delta_N \in \mathbb{R}^{L\times L}$ denote the discrete Neumann Laplacian (Eq.~\eqref{eq:neumann_laplacian}), and define the single-step diffusion operator
\begin{equation}
\mathcal{S}_\alpha \triangleq I + \alpha \Delta_N, \qquad 0 \le \alpha < \tfrac{1}{2},
\end{equation}
applied channel-wise to $X\in\mathbb{R}^{L\times d}$.

\begin{lemma}[Spectrum of the Discrete Neumann Laplacian]
\label{lem:laplacian_spectrum}
The matrix $\Delta_N$ is real symmetric and negative semi-definite. Its eigenvalues are
\begin{equation}
\lambda_k = -4\sin^2\!\left(\frac{\pi k}{2L}\right), \quad k=0,1,\ldots,L-1,
\label{eq:laplacian_eigenvalues}
\end{equation}
where $\lambda_0=0$ (constant mode) and $\lambda_k\in[-4,0]$ for all $k$.
\end{lemma}

\begin{theorem}[Spectral Non-Expansiveness and Strict Contraction]
\label{thm:nonexpansive}
For any $X\in\mathbb{R}^{L\times d}$ and $0\le\alpha<\tfrac{1}{2}$,
\begin{equation}
\lVert\mathcal{S}_\alpha\rVert_2 \le 1,
\qquad
\lVert\mathcal{S}_\alpha(X)\rVert_F \le \lVert X\rVert_F.
\label{eq:spectral_nonexpansive}
\end{equation}
Let $\Pi_\perp$ be the orthogonal projection onto $\mathrm{Null}(\Delta_N)^\perp$ (equivalently, the zero-mean subspace along the sequence dimension). Then
\begin{equation}
\begin{aligned}
\lVert\Pi_\perp \mathcal{S}_\alpha(X)\rVert_F
&\le \kappa(\alpha,L)\,\lVert\Pi_\perp X\rVert_F,\\
\kappa(\alpha,L) &= \max_{k\ge 1}\bigl(1+\alpha\lambda_k\bigr) < 1.
\end{aligned}
\label{eq:strict_contraction}
\end{equation}
\end{theorem}

The constraint $\alpha<\tfrac{1}{2}$ is a discrete stability budget analogous to the CFL condition for explicit diffusion: it ensures non-expansiveness and strict attenuation of non-constant (high-frequency) components, while leaving the constant mode unchanged.

To quantify locality, we use the discrete Dirichlet energy.

\begin{definition}[Discrete Dirichlet Energy]
\label{def:dirichlet_energy}
For $X\in\mathbb{R}^{L\times d}$,
\begin{equation}
\mathcal{E}(X)
\triangleq
\frac{1}{2}\langle X,-\Delta_N X\rangle
=
\frac{1}{2}\sum_{j=1}^d \sum_{i=1}^{L-1} (X_{i+1,j}-X_{i,j})^2,
\label{eq:dirichlet_energy}
\end{equation}
which measures local roughness along the sequence axis.
\end{definition}

\begin{theorem}[Monotone Energy Dissipation]
\label{thm:energy_dissipation}
For any $0\le\alpha<\tfrac{1}{2}$ and $X\in\mathbb{R}^{L\times d}$,
\begin{equation}
\mathcal{E}(\mathcal{S}_\alpha(X)) \le \mathcal{E}(X),
\label{eq:energy_monotone}
\end{equation}
with strict inequality whenever $\Pi_\perp X\neq 0$ and $\alpha>0$.
\end{theorem}

\begin{corollary}[Multi-Scale Stability Under a Global Budget]
\label{cor:multiscale_stability}
Let $\mathcal{S}_{\boldsymbol{\alpha}}$ be the multi-scale diffusion operator defined above. If $\sum_{k=1}^K \alpha_k < \tfrac{1}{2}$, then $\mathcal{S}_{\boldsymbol{\alpha}}$ is spectrally non-expansive and dissipates $\mathcal{E}(\cdot)$ monotonically.
\end{corollary}

\paragraph{Practical takeaway.}
With a strict global budget ($\alpha<\tfrac12$ or $\sum_k\alpha_k<\tfrac12$), PDE diffusion is a safe architectural primitive: it cannot amplify representations and it monotonically reduces local roughness.

\subsection{Layer II: Compositional Perturbation Bounds}
\label{subsec:compositional_bounds}

We next study how diffusion perturbs a Transformer computation when inserted at different locations. Let $f_\ell:\mathbb{R}^{L\times d}\to\mathbb{R}^{L\times d}$ denote sub-modules (LayerNorm, attention, MLP), and consider the composition $F=f_m\circ\cdots\circ f_1$.

\begin{assumption}[Local Lipschitz Regularity]
\label{assump:lipschitz}
For each $\ell$, there exists $L_\ell>0$ such that in a neighborhood of interest,
\begin{equation}
\lVert f_\ell(X)-f_\ell(Y)\rVert_F \le L_\ell \lVert X-Y\rVert_F.
\label{eq:lipschitz_assumption}
\end{equation}
\end{assumption}

Define the variant with diffusion inserted after $f_j$:
\begin{equation}
F^{(j)} \triangleq f_m \circ \cdots \circ f_{j+1} \circ \mathcal{S}_\alpha \circ f_j \circ \cdots \circ f_1.
\label{eq:insertion_composition}
\end{equation}

\begin{theorem}[Perturbation Propagation Bound]
\label{thm:lipschitz_bound}
Under Assumption~\ref{assump:lipschitz}, for any input $X$ and $Z_j=(f_j\circ\cdots\circ f_1)(X)$,
\begin{equation}
\begin{aligned}
\lVert F^{(j)}(X)-F(X)\rVert_F
&\le
\left(\prod_{\ell=j+1}^m L_\ell\right)\alpha\,\lVert\Delta_N Z_j\rVert_F\\
&\le
\left(\prod_{\ell=j+1}^m L_\ell\right)\alpha\,\sqrt{2\mathcal{E}(Z_j)}.
\end{aligned}
\label{eq:lipschitz_propagation}
\end{equation}
\end{theorem}

This bound isolates two insertion-dependent quantities: the local roughness at the insertion point, captured by $\mathcal{E}(Z_j)$, and the downstream sensitivity, captured by $\prod_{\ell=j+1}^m L_\ell$. In particular, diffusion is guaranteed to be a small perturbation whenever both $\mathcal{E}(Z_j)$ and the downstream amplification are controlled; conversely, sensitivity to insertion position is expected when either factor is large.

\paragraph{Practical takeaway.}
The same diffusion operator can help or hurt depending on where it is inserted: its effect scales with insertion-point roughness and downstream amplification, explaining why early insertion is typically most beneficial.

\subsection{Layer III: Diffusion--Attention Non-Commutativity as a Diagnostic}
\label{subsec:noncommutativity}

Finally, we formalize the difference between pre-attention and post-attention diffusion through an exact commutator identity. Let $\mathcal{A}$ denote the (possibly input-dependent) self-attention operator.

\begin{proposition}[Commutator Identity]
\label{prop:commutator_identity}
For any $X$, define
\begin{equation}
\begin{aligned}
Y_{\mathrm{pre}} &\triangleq \mathcal{A}(\mathcal{S}_\alpha(X)),\\
Y_{\mathrm{post}} &\triangleq \mathcal{S}_\alpha(\mathcal{A}(X)).
\end{aligned}
\label{eq:pre_post_def}
\end{equation}
Then
\begin{equation}
\begin{aligned}
Y_{\mathrm{post}}-Y_{\mathrm{pre}}
&=
[\mathcal{S}_\alpha,\mathcal{A}]X\\
&\triangleq
\mathcal{S}_\alpha(\mathcal{A}(X)) - \mathcal{A}(\mathcal{S}_\alpha(X)),
\end{aligned}
\label{eq:commutator_diff}
\end{equation}
and hence $\lVert Y_{\mathrm{post}}-Y_{\mathrm{pre}}\rVert_F=\lVert[\mathcal{S}_\alpha,\mathcal{A}]X\rVert_F$.
\end{proposition}

To interpret this quantity, we use a local linearization at a fixed input $X$: approximate attention as a mixing operator along the sequence dimension, $\mathcal{A}(X)\approx A(X)X$, where $A(X)\in\mathbb{R}^{L\times L}$ is row-stochastic. Substituting into the commutator yields the heuristic form
\begin{equation}
[\mathcal{S}_\alpha,\mathcal{A}]X
\approx
\alpha\bigl(\Delta_N A(X)-A(X)\Delta_N\bigr)X,
\label{eq:commutator_linearized}
\end{equation}
which makes explicit that non-commutativity arises from a mismatch between local geometry ($\Delta_N$) and non-local routing ($A(X)$). We emphasize that Eq.~\eqref{eq:commutator_linearized} is a diagnostic approximation rather than a global guarantee: self-attention is nonlinear and input-dependent, so this layer provides interpretable guidance, not a universal bound.

\paragraph{Practical takeaway.}
Large diffusion--attention mismatch (a large commutator) indicates structural conflict between local smoothing and non-local routing, and motivates avoiding post-attention diffusion when attention induces strong non-local mixing.

\subsection{Design Implications}
\label{subsec:design_guidelines}

The analysis suggests three practical implications. First, PDE diffusion provides a certified local geometric prior (Layer~I) that is stable under a strict budget $\alpha<\tfrac{1}{2}$. Second, the effect of inserting PDE diffusion is controlled by insertion-point roughness and downstream amplification (Layer~II), giving a principled notion of ``position sensitivity'' without relying on empirical observations. Third, the commutator $[\mathcal{S}_\alpha,\mathcal{A}]$ exactly measures the representational discrepancy between pre- and post-attention diffusion, and its linearized form attributes large discrepancies to incompatible local and non-local mixing patterns (Layer~III). Together, these statements justify preferring PDE diffusion placements that regularize representations before aggressive non-local routing, while treating post-routing smoothing as potentially disruptive when the induced commutator is large.

\section{Experiments}
\label{sec:experiments}

This section empirically validates the core claims of our operator-theoretic framework. We first compare insertion positions to test the placement predictions implied by the compositional and non-commutativity analyses (Section~\ref{sec:theory}). We then ablate the multi-scale design under a fixed global stability budget to test the frequency-coverage hypothesis. All experiments are conducted on a challenging long-sequence benchmark to provide a stringent test of the theory.

\subsection{Experimental Setup}

We conduct all experiments on the \textbf{Long Range Arena (LRA)} benchmark \cite{tay2020long,tay2021longrange}, which includes five tasks spanning diverse modalities and challenges. Our models integrate the proposed PDE diffusion layer into a strong, optimized vanilla Transformer baseline, following common architectural and training practices \cite{devlin2019bert,liu2019roberta,raffel2020exploring}. We systematically evaluate the seven canonical insertion positions defined in Table~\ref{tab:positions} to determine where locality should be introduced. All models were trained on NVIDIA A100-80GB GPUs, with reproducibility ensured by fixing random seeds. Additional dataset, configuration, and training details are provided in Appendix~\ref{sec:supplementary}.

\subsection{Validating Position-Dependent Predictions}

Before presenting the results, we clarify how our experiments are designed to directly test the three-layer operator-theoretic framework introduced in Section~\ref{sec:theory}. 
We reuse the canonical diffusion insertion strategies defined in Table~\ref{tab:positions}, spanning early (After-Embedding) to late (After-Attention) placements along with several intermediate variants.

\paragraph{Tested predictions.}
Our analysis yields three simple, falsifiable placement predictions. (\textbf{Pred-1}) Early PDE diffusion (e.g., After-Embedding) should be most beneficial, since representations are rough and regularization has maximal leverage (Layer~II). (\textbf{Pred-2}) Post-attention PDE diffusion should be harmful due to a structural double-mixing conflict between local smoothing and non-local routing (Layer~III). (\textbf{Pred-3}) Intermediate placements should yield smaller gains as representations become progressively smoother through depth (Layer~II).

\paragraph{Main results.}
Across five Long Range Arena tasks, the results strongly validate the position-dependent predictions of our framework. Consistent with \textbf{Pred-1} (Layer~II), early PDE diffusion yields the best performance: inserting PDE diffusion immediately after the embedding layer (P1) improves average accuracy from $58.62\%$ to $62.69\%$ ($+4.1$ points). Consistent with \textbf{Pred-2} (Layer~III), post-attention PDE diffusion is harmful: After-Attention insertion (P7) drops to $56.17\%$ ($-2.5$ points), matching the predicted double-mixing conflict between local smoothing and non-local routing. Intermediate insertion strategies (P2--P4) produce smaller gains, consistent with \textbf{Pred-3} (Layer~II) and the diminishing-leverage interpretation as representations become smoother through depth. Overall, insertion position is not a tunable detail but a theoretically constrained design choice governed by local--global operator interaction.

\begin{table}[H]
\centering
\caption{Average accuracy on the LRA benchmark for different PDE integration positions. Detailed per-task scores are in Appendix B.}
\label{tab:main_results_summary}
\begin{tabular*}{\columnwidth}{@{\extracolsep{\fill}}lc}
\toprule
\textbf{Integration Position} & \textbf{Avg. Accuracy} \\
\midrule
\textbf{PDE-After-Embedding} & \textbf{0.6269} \\
PDE-After-MLP & 0.5986 \\
PDE-Layer-Diffusion & 0.5970 \\
PDE-Before-LayerNorm & 0.5962 \\
PDE-In-Attention & 0.5909 \\
PDE-Head-Diffusion & 0.5884 \\
\midrule
\textbf{Baseline Transformer} & \textbf{0.5862} \\
\midrule
PDE-After-Attention & 0.5617 \\
\bottomrule
\end{tabular*}
\end{table}

\begin{figure*}[t]
  \centering
  \includegraphics[width=\textwidth]{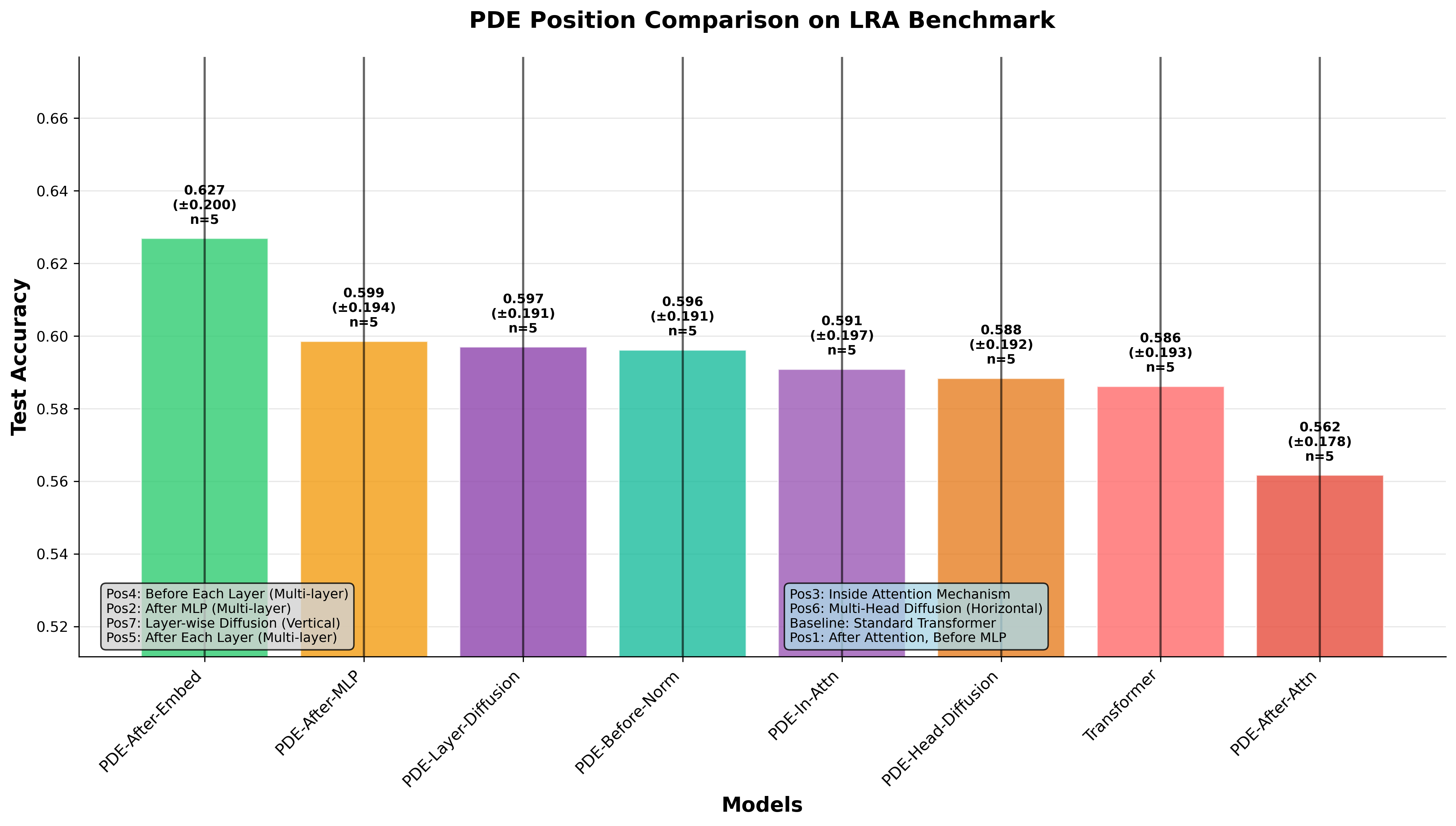}
  \caption{Overall performance comparison on the LRA benchmark. Error
    bars show standard deviation across five runs ($n{=}5$).}
  \label{fig:main_results_comparison}
\end{figure*}

\subsection{Validating Multi-Scale Frequency Coverage}

We next evaluate the prediction of Corollary~\ref{cor:multiscale_stability}, which states that multi-scale PDE diffusion preserves intrinsic stability under the global budget constraint $\sum_k \alpha_k < 1/2$ while providing broader frequency coverage than any single scale. 
To test this claim, we focus on the ListOps task and compare three single-scale variants (Fast $h{=}1$, Medium $h{=}2$, Slow $h{=}4$) against an adaptive multi-scale configuration combining $h{=}1,2,4$, evaluated at the top three insertion positions. 
All settings enforce the same total budget $\sum_k \alpha_k = 0.48 < 1/2$ to isolate the effect of frequency coverage.

As shown in Table~\ref{tab:multiscale_complete_appendix} (Appendix), adaptive multi-scale diffusion consistently outperforms the best single-scale baseline across all positions, yielding improvements of $+0.90$ percentage points for After-Embedding, $+0.80$ for After-MLP, and $+0.60$ for Layer-Diffusion. 
These gains support the theoretical prediction that different diffusion scales capture complementary frequency bands (see Appendix~C.3), and that their combination provides more uniform spectral coverage without violating stability guarantees. 
The consistency of improvements across insertion positions suggests that frequency heterogeneity is a general characteristic of Long Range Arena tasks, reinforcing multi-scale diffusion as a robust design principle rather than a position-specific optimization.

\begin{figure*}[t]
  \centering
  \includegraphics[width=\textwidth]{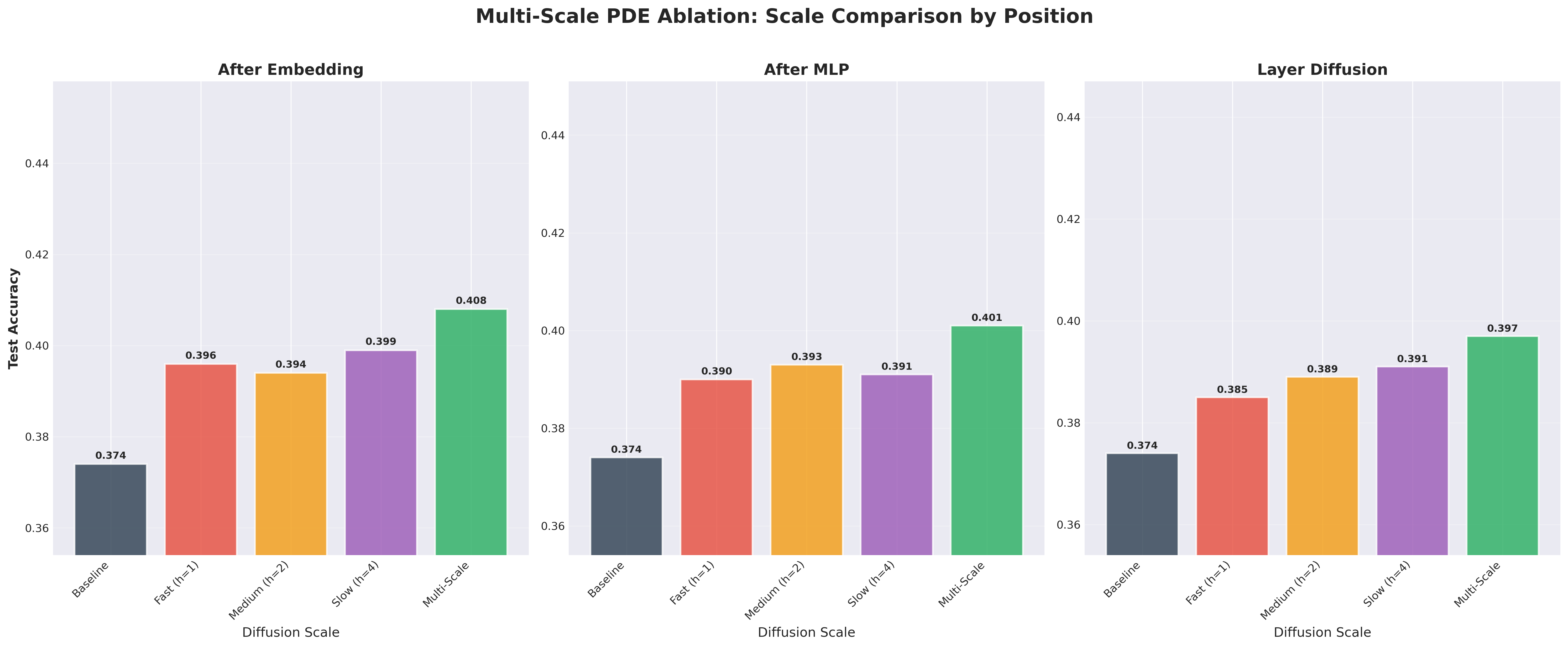}
  \caption{Multi-scale ablation on ListOps. The adaptive multi-scale configuration consistently outperforms all single-scale variants, validating Corollary~\ref{cor:multiscale_stability} under a fixed global stability budget.}
  \label{fig:multiscale_breakdown}
\end{figure*}

\subsection{Conclusion of Experiments}

Our experiments on Long Range Arena \cite{tay2020long,tay2021longrange} provide a controlled validation of the proposed operator-theoretic framework across all three analytical layers. 
First, consistent and stable training under $\alpha = 0.1 < 1/2$ confirms that the spectral budget from Layer~I (intrinsic stability) is practically viable. 
Second, position-dependent comparisons (Table~\ref{tab:detailed_per_task_scores} in the appendix) align closely with the compositional bound in Theorem~\ref{thm:lipschitz_bound}: early-stage insertion (P1) yields a $+4.1$ percentage point improvement, while late-stage insertion after attention (P7) results in a $-2.5$ point degradation, supporting the predicted roughness--sensitivity trade-off from Layer~II. 
Third, the systematic degradation observed at P7 is consistent with the Layer~III non-commutativity analysis, lending empirical support to the double-mixing hypothesis implied by the commutator $[\mathcal{S}, \mathcal{A}]$, although we do not directly measure commutator norms in this work. 
Finally, the multi-scale extension predicted by Corollary~\ref{cor:multiscale_stability} is supported by consistent gains in the ablation study (Table~\ref{tab:multiscale_complete_appendix} in the appendix). 
At the same time, this empirical validation is intentionally scoped: LRA is a controlled benchmark rather than a large-scale language modeling setting \cite{brown2020language}; we treat $\mathcal{E}(Z_j)$ and $[\mathcal{S},\mathcal{A}]$ as explanatory quantities rather than directly measured causal signals; and multi-scale ablations are limited to a single task (ListOps). Overall, the evidence supports the main claim that operator-theoretic reasoning can predict insertion-position sensitivity, while leaving broader scaling and mechanistic measurement as open directions.

\section{Conclusion}
\label{sec:conclusion}

We introduced an operator-theoretic framework for understanding \emph{where} to add locality in Transformers. Under a simple stability budget ($\alpha<1/2$), a PDE diffusion layer provides a certified non-expansive local prior; when composed with the Transformer backbone, its impact becomes position-dependent via a roughness--sensitivity trade-off; and when composed with attention, non-commutativity explains why post-attention smoothing can be structurally harmful.

On Long Range Arena \cite{tay2020long,tay2021longrange}, these predictions are borne out: inserting PDE diffusion after embedding improves average accuracy by $+4.1$ points, while inserting PDE diffusion after attention degrades performance by $-2.5$ points, and a multi-scale PDE variant yields consistent gains under the same global budget. More broadly, the main contribution is a transferable template for reasoning about local--global compositions, separating provable guarantees from compositional bounds and diagnostic heuristics.

\paragraph{Scope and limitations.}
Our claims are scoped to (i) a spectrally-budgeted PDE diffusion operator along the sequence dimension and (ii) controlled long-sequence benchmarks (LRA) with moderate model sizes and contexts. We do not directly measure geometric diagnostics (e.g., commutator norms) in the current study, and the multi-scale ablation is limited in task coverage.

\paragraph{Future work.}
Promising directions include mechanistic validation by measuring roughness and commutator statistics across depth; scaling to language modeling benchmarks and larger models; integrating PDE diffusion more tightly inside attention computations; and applying the same operator-theoretic lens to other hybrid local--global modules (e.g., convolution--attention and state-space--attention).

\section{Impact Statements}
 Our research does not involve human subjects, personally identifiable information, or sensitive data. All datasets used in experiments are publicly available and have been widely adopted in prior work. We are not aware of any foreseeable misuse or harmful applications directly stemming from our proposed methods. We further confirm that there are no conflicts of interest, sponsorship concerns, or legal compliance issues associated with this work. We acknowledge that any potential societal impact of large-scale models, such as fairness, bias, or privacy concerns, lies beyond the specific scope of this paper but remains an important consideration for future research.

\bibliographystyle{icml2026}
\bibliography{example_paper}

\begin{thebibliography}{28}
\providecommand{\natexlab}[1]{#1}
\providecommand{\url}[1]{\texttt{#1}}
\expandafter\ifx\csname urlstyle\endcsname\relax
  \providecommand{\doi}[1]{doi: #1}\else
  \providecommand{\doi}{doi: \begingroup \urlstyle{rm}\Url}\fi

\bibitem[Ba et~al.(2016)Ba, Kiros, and Hinton]{ba2021layer}
Ba, J.~L., Kiros, J.~R., and Hinton, G.~E.
\newblock Layer normalization.
\newblock In \emph{arXiv preprint arXiv:1607.06450}, 2016.
\newblock URL \url{https://arxiv.org/abs/1607.06450}.

\bibitem[Beltagy et~al.(2020)Beltagy, Peters, and Cohan]{beltagy2020longformer}
Beltagy, I., Peters, M.~E., and Cohan, A.
\newblock Longformer: The long-document transformer.
\newblock In \emph{Proceedings of the 2020 Conference on Empirical Methods in Natural Language Processing (EMNLP)}, 2020.
\newblock \doi{10.18653/v1/2020.emnlp-main.19}.
\newblock URL \url{https://arxiv.org/abs/2004.05150}.

\bibitem[Brown et~al.(2020)Brown, Mann, Ryder, Subbiah, Kaplan, Dhariwal, Neelakantan, Shyam, Sastry, Askell, Agarwal, Herbert-Voss, Krueger, Henighan, Child, Ramesh, Ziegler, Wu, Winter, Hesse, Chen, Sigler, Litwin, Gray, Chess, Clark, Berner, McCandlish, Radford, Sutskever, and Amodei]{brown2020language}
Brown, T., Mann, B., Ryder, N., Subbiah, M., Kaplan, J.~D., Dhariwal, P., Neelakantan, A., Shyam, P., Sastry, G., Askell, A., Agarwal, S., Herbert-Voss, A., Krueger, G., Henighan, T., Child, R., Ramesh, A., Ziegler, D., Wu, J., Winter, C., Hesse, C., Chen, M., Sigler, E., Litwin, M., Gray, S., Chess, B., Clark, J., Berner, C., McCandlish, S., Radford, A., Sutskever, I., and Amodei, D.
\newblock Language models are few-shot learners.
\newblock In \emph{Advances in Neural Information Processing Systems}, volume~33, 2020.
\newblock URL \url{https://arxiv.org/abs/2005.14165}.

\bibitem[Child et~al.(2019)Child, Gray, Radford, and Sutskever]{child2019sparse}
Child, R., Gray, S., Radford, A., and Sutskever, I.
\newblock Generating long sequences with sparse transformers.
\newblock In \emph{arXiv preprint arXiv:1904.10509}, 2019.
\newblock URL \url{https://arxiv.org/abs/1904.10509}.

\bibitem[Choromanski et~al.(2020)Choromanski, Likhosherstov, Dohan, Song, Gane, Sarlos, Hawkins, Davis, Mohiuddin, Kaiser, Belanger, Colwell, and Weller]{choromanski2021rethinking}
Choromanski, K., Likhosherstov, V., Dohan, D., Song, X., Gane, A., Sarlos, T., Hawkins, P., Davis, J., Mohiuddin, A., Kaiser, L., Belanger, D., Colwell, L., and Weller, A.
\newblock Rethinking attention with performers.
\newblock In \emph{International Conference on Learning Representations}, 2020.
\newblock URL \url{https://arxiv.org/abs/2009.14794}.
\newblock Published at ICLR 2021.

\bibitem[Dai et~al.(2020)Dai, Lai, Yang, and Le]{dai2020funnel}
Dai, Z., Lai, G., Yang, Y., and Le, Q.~V.
\newblock Funnel-transformer: Filtering out sequential redundancy for efficient language processing.
\newblock In \emph{Advances in Neural Information Processing Systems}, volume~33, 2020.
\newblock URL \url{https://arxiv.org/abs/2006.03236}.

\bibitem[Dao et~al.(2022)Dao, Fu, Ermon, Rudra, and R{\'e}]{dao2022flashattention}
Dao, T., Fu, D.~Y., Ermon, S., Rudra, A., and R{\'e}, C.
\newblock Flashattention: Fast and memory-efficient exact attention with io-awareness.
\newblock In \emph{Advances in Neural Information Processing Systems}, volume~35, 2022.
\newblock URL \url{https://arxiv.org/abs/2205.14135}.

\bibitem[Devlin et~al.(2018)Devlin, Chang, Lee, and Toutanova]{devlin2019bert}
Devlin, J., Chang, M.-W., Lee, K., and Toutanova, K.
\newblock {BERT}: Pre-training of deep bidirectional transformers for language understanding.
\newblock In \emph{Proceedings of the 2019 Conference of the North American Chapter of the Association for Computational Linguistics: Human Language Technologies}, volume~1, pp.\  4171--4186, 2018.
\newblock \doi{10.18653/v1/N19-1423}.
\newblock URL \url{https://arxiv.org/abs/1810.04805}.
\newblock Published at NAACL 2019.

\bibitem[Gu \& Dao(2023)Gu and Dao]{gu2023mamba}
Gu, A. and Dao, T.
\newblock Mamba: Linear-time sequence modeling with selective state spaces.
\newblock In \emph{Advances in Neural Information Processing Systems}, volume~36, 2023.
\newblock URL \url{https://arxiv.org/abs/2312.00752}.

\bibitem[Gu et~al.(2021)Gu, Goel, and R{\'e}]{gu2022s4}
Gu, A., Goel, K., and R{\'e}, C.
\newblock Efficiently modeling long sequences with structured state spaces.
\newblock In \emph{International Conference on Learning Representations}, 2021.
\newblock URL \url{https://arxiv.org/abs/2111.00396}.
\newblock Published at ICLR 2022.

\bibitem[Haber \& Ruthotto(2017)Haber and Ruthotto]{haber2017stable}
Haber, E. and Ruthotto, L.
\newblock Stable architectures for deep neural networks.
\newblock \emph{Inverse Problems}, 34\penalty0 (1):\penalty0 014004, 2017.
\newblock \doi{10.1088/1361-6420/aa9a90}.
\newblock URL \url{https://arxiv.org/abs/1705.03341}.
\newblock Published in Inverse Problems 2018.

\bibitem[He et~al.(2015)He, Zhang, Ren, and Sun]{he2016deep}
He, K., Zhang, X., Ren, S., and Sun, J.
\newblock Deep residual learning for image recognition.
\newblock In \emph{Proceedings of the IEEE Conference on Computer Vision and Pattern Recognition}, pp.\  770--778, 2015.
\newblock \doi{10.1109/CVPR.2016.90}.
\newblock URL \url{https://arxiv.org/abs/1512.03385}.
\newblock Published at CVPR 2016.

\bibitem[Ho et~al.(2020)Ho, Jain, and Abbeel]{ho2020denoising}
Ho, J., Jain, A., and Abbeel, P.
\newblock Denoising diffusion probabilistic models.
\newblock In \emph{Advances in Neural Information Processing Systems}, volume~33, 2020.
\newblock URL \url{https://arxiv.org/abs/2006.11239}.

\bibitem[Katharopoulos et~al.(2020)Katharopoulos, Vyas, Pappas, and Fleuret]{katharopoulos2020transformers}
Katharopoulos, A., Vyas, A., Pappas, N., and Fleuret, F.
\newblock Transformers are rnns: Fast autoregressive transformers with linear attention.
\newblock In \emph{Proceedings of the 37th International Conference on Machine Learning}, volume 119, pp.\  5156--5165, 2020.
\newblock URL \url{https://arxiv.org/abs/2006.16236}.

\bibitem[Kitaev et~al.(2020)Kitaev, Kaiser, and Levskaya]{kitaev2020reformer}
Kitaev, N., Kaiser, L., and Levskaya, A.
\newblock Reformer: The efficient transformer.
\newblock In \emph{International Conference on Learning Representations}, 2020.
\newblock URL \url{https://arxiv.org/abs/2001.04451}.

\bibitem[Li et~al.(2020)Li, Kovachki, Azizzadenesheli, Liu, Bhattacharya, Stuart, and Anandkumar]{li2021fourier}
Li, Z., Kovachki, N.~B., Azizzadenesheli, K., Liu, B., Bhattacharya, K., Stuart, A., and Anandkumar, A.
\newblock Fourier neural operator for parametric partial differential equations.
\newblock In \emph{International Conference on Learning Representations}, 2020.
\newblock URL \url{https://arxiv.org/abs/2010.08895}.
\newblock Published at ICLR 2021.

\bibitem[Liu et~al.(2019)Liu, Ott, Goyal, Du, Joshi, Chen, Levy, Lewis, Zettlemoyer, and Stoyanov]{liu2019roberta}
Liu, Y., Ott, M., Goyal, N., Du, J., Joshi, M., Chen, D., Levy, O., Lewis, M., Zettlemoyer, L., and Stoyanov, V.
\newblock {RoBERTa}: A robustly optimized {BERT} pretraining approach.
\newblock In \emph{arXiv preprint arXiv:1907.11692}, 2019.
\newblock URL \url{https://arxiv.org/abs/1907.11692}.

\bibitem[Miyato et~al.(2018)Miyato, Kataoka, Koyama, and Yoshida]{miyato2018spectral}
Miyato, T., Kataoka, T., Koyama, M., and Yoshida, Y.
\newblock Spectral normalization for generative adversarial networks.
\newblock In \emph{International Conference on Learning Representations}, 2018.
\newblock URL \url{https://arxiv.org/abs/1802.05957}.

\bibitem[Raffel et~al.(2019)Raffel, Shazeer, Roberts, Lee, Narang, Matena, Zhou, Li, and Liu]{raffel2020exploring}
Raffel, C., Shazeer, N., Roberts, A., Lee, K., Narang, S., Matena, M., Zhou, Y., Li, W., and Liu, P.~J.
\newblock Exploring the limits of transfer learning with a unified text-to-text transformer.
\newblock volume~21, pp.\  1--67, 2019.
\newblock URL \url{https://arxiv.org/abs/1910.10683}.
\newblock Published in JMLR 2020.

\bibitem[Ruthotto \& Haber(2018)Ruthotto and Haber]{ruthotto2017deep}
Ruthotto, L. and Haber, E.
\newblock Deep neural networks motivated by partial differential equations.
\newblock \emph{Journal of Mathematical Imaging and Vision}, 62\penalty0 (3):\penalty0 352--364, 2018.
\newblock \doi{10.1007/s10851-019-00903-1}.
\newblock URL \url{https://arxiv.org/abs/1804.04272}.
\newblock Published in Journal of Mathematical Imaging and Vision 2020.

\bibitem[Tay et~al.(2020{\natexlab{a}})Tay, Dehghani, Abnar, Shen, Bahri, Pham, Rao, Yang, Ruder, and Metzler]{tay2021longrange}
Tay, Y., Dehghani, M., Abnar, S., Shen, Y., Bahri, D., Pham, P., Rao, J., Yang, L., Ruder, S., and Metzler, D.
\newblock Long range arena: A benchmark for efficient transformers.
\newblock In \emph{International Conference on Learning Representations}, 2020{\natexlab{a}}.
\newblock URL \url{https://arxiv.org/abs/2011.04006}.
\newblock Published at ICLR 2021.

\bibitem[Tay et~al.(2020{\natexlab{b}})Tay, Dehghani, Bahri, and Metzler]{tay2020long}
Tay, Y., Dehghani, M., Bahri, D., and Metzler, D.
\newblock Efficient transformers: A survey.
\newblock \emph{ACM Computing Surveys}, 55\penalty0 (6):\penalty0 109:1--109:28, 2020{\natexlab{b}}.
\newblock \doi{10.1145/3530811}.
\newblock URL \url{https://arxiv.org/abs/2009.06732}.
\newblock Published in ACM Computing Surveys 2022.

\bibitem[Vaswani et~al.(2017)Vaswani, Shazeer, Parmar, Uszkoreit, Jones, Gomez, Kaiser, and Polosukhin]{vaswani2017attention}
Vaswani, A., Shazeer, N., Parmar, N., Uszkoreit, J., Jones, L., Gomez, A.~N., Kaiser, {\L}., and Polosukhin, I.
\newblock Attention is all you need.
\newblock In \emph{Advances in Neural Information Processing Systems}, volume~30, 2017.
\newblock URL \url{https://arxiv.org/abs/1706.03762}.

\bibitem[Wang et~al.(2020)Wang, Li, Khabsa, Fang, and Ma]{wang2020linformer}
Wang, S., Li, B.~Z., Khabsa, M., Fang, H., and Ma, H.
\newblock Linformer: Self-attention with linear complexity.
\newblock In \emph{Advances in Neural Information Processing Systems}, volume~33, 2020.
\newblock URL \url{https://arxiv.org/abs/2006.04768}.

\bibitem[Weickert(1998)]{weickert1998anisotropic}
Weickert, J.
\newblock \emph{Anisotropic Diffusion in Image Processing}.
\newblock Teubner, 1998.
\newblock ISBN 978-3-519-02606-9.

\bibitem[Wu et~al.(2019)Wu, Fan, Baevski, Dauphin, and Auli]{wu2019pay}
Wu, F., Fan, A., Baevski, A., Dauphin, Y., and Auli, M.
\newblock Pay less attention with lightweight and dynamic convolutions.
\newblock In \emph{International Conference on Learning Representations}, 2019.
\newblock URL \url{https://arxiv.org/abs/1901.10430}.

\bibitem[Xiong et~al.(2021)Xiong, Zeng, Chakraborty, Tan, Fung, Li, and Singh]{xiong2021nystromformer}
Xiong, Y., Zeng, Z., Chakraborty, R., Tan, M., Fung, G., Li, Y., and Singh, V.
\newblock Nystr{\"o}mformer: A nystr{\"o}m-based algorithm for approximating self-attention.
\newblock In \emph{Proceedings of the AAAI Conference on Artificial Intelligence}, volume~35, pp.\  14138--14148, 2021.
\newblock \doi{10.1609/aaai.v35i16.17664}.
\newblock URL \url{https://arxiv.org/abs/2102.03902}.

\bibitem[Zaheer et~al.(2020)Zaheer, Guruganesh, Dubey, Ainslie, Alberti, Onta{\~n}{\'o}n, Pham, Ravula, Wang, Yang, and Ahmed]{zaheer2020bigbird}
Zaheer, M., Guruganesh, G., Dubey, K.~A., Ainslie, J., Alberti, C., Onta{\~n}{\'o}n, S., Pham, P., Ravula, A., Wang, Q., Yang, L., and Ahmed, A.
\newblock Big bird: Transformers for longer sequences.
\newblock In \emph{Advances in Neural Information Processing Systems}, volume~33, 2020.
\newblock URL \url{https://arxiv.org/abs/2007.14062}.

\end{thebibliography}

\newpage
\appendix
\onecolumn
\section*{Reproducibility Statement}
We place strong emphasis on the reproducibility of our results. In the main text, we clearly describe our proposed theoretical framework, its derivation, and the algorithmic implementation details. A complete set of proofs for theoretical results is provided in the Appendix. Hyperparameter configurations, training setups, and ablation study details are explicitly reported in the Experiments section. All datasets used are standard and publicly available, and their preprocessing steps are fully documented in the supplementary material. To facilitate replication, we will release anonymized source code and scripts for reproducing all experiments upon publication.

\section*{LLM Usage Statement}
In accordance with ICML guidelines on the disclosure of Large Language Model (LLM) usage, we clarify that no LLM contributed substantively to the conception, methodology, or analysis presented in this paper. LLMs (e.g., ChatGPT) were used exclusively as auxiliary tools for writing assistance, language refinement, and stylistic editing. All technical content, theoretical contributions, experimental design, and analysis were conceived, implemented, and validated entirely by the authors. The role of LLMs was limited to improving clarity of presentation and does not rise to the level of authorship or contribution under ICML policy.

\section{Supplementary Material}
\label{sec:supplementary}

This appendix provides supplementary material to support the main paper, including experimental setup details, extended results, and additional analyses.

\subsection{Detailed Experimental Setup}

\subsubsection{Datasets and Tasks.}
We evaluate our PDE-enhanced Transformer models on the Long Range Arena (LRA) benchmark \cite{tay2020long,tay2021longrange}, which consists of five challenging tasks designed to assess the ability of sequence models to capture long-range dependencies:
\begin{itemize}
    \item \textbf{ListOps}: A synthetic task requiring hierarchical reasoning over sequences of up to 2,000 tokens, with 10 classes.
    \item \textbf{Text Classification}: Document classification on IMDb movie reviews with sequences up to 4,000 tokens and 2 classes.
    \item \textbf{Text Retrieval}: Matching queries with relevant documents, with sequences up to 4,000 tokens and 2 classes.
    \item \textbf{PathFinder}: A visual reasoning task on 32$\times$32 images involving path connectivity detection, flattened to 1,024-token sequences, with 2 classes.
    \item \textbf{Image Classification}: CIFAR-10 classification on 32$\times$32 images, flattened to 1,024-token sequences, with 10 classes.
\end{itemize}

\subsubsection{Software and Hardware Environment.}
All experiments were conducted on a high-performance computing cluster equipped with NVIDIA A100-80GB GPUs. The software stack included PyTorch 1.12.1, CUDA 11.3, and Python 3.9. Reproducibility was ensured by fixing the random seeds for Python (42), NumPy (0), and PyTorch (1) across all runs.

\subsubsection{Model Hyperparameters.}
All experiments use Transformer configurations optimized for A100-80GB GPUs. Configurations were tuned per task to maximize GPU utilization while maintaining training stability, as detailed in Table~\ref{tab:model_configs}. A unified set of training hyperparameters, shown in Table~\ref{tab:training_hyperparams}, was used across all runs to ensure fair comparisons.

\begin{table}[H]
\centering
\caption{Task-specific model configurations.}
\label{tab:model_configs}
\begin{tabular*}{\columnwidth}{@{\extracolsep{\fill}}lcccc}
\toprule
\textbf{Task} & \textbf{Dim} & \textbf{Layers} & \textbf{Heads} & \textbf{Batch Size} \\
\midrule
ListOps & 128 & 6 & 8 & 256 \\
PathFinder & 128 & 6 & 8 & 512 \\
Text Cls. & 128 & 6 & 8 & 128 \\
Text Retrieval & 128 & 4 & 8 & 64 \\
Image Cls. & 128 & 4 & 8 & 1024 \\
\bottomrule
\end{tabular*}
\end{table}

\begin{table}[H]
\centering
\caption{Unified training hyperparameters.}
\label{tab:training_hyperparams}
\begin{tabular*}{\columnwidth}{@{\extracolsep{\fill}}ll}
\toprule
\textbf{Hyperparameter} & \textbf{Value} \\
\midrule
Optimizer & AdamW ($\beta_1=0.9, \beta_2=0.98$) \\
Learning Rate & $10^{-3}$ with linear warmup \\
& (10,000 steps) \& cosine decay \\
Training Epochs & 50 with early stopping (patience=10) \\
Weight Decay & $10^{-5}$ \\
Gradient Clipping & Max norm 1.0 \\
Dropout & 0.1 \\
MLP Hidden Dim & 512 (4x model dimension) \\
\bottomrule
\end{tabular*}
\end{table}
\subsubsection{The detailed PDE operation}
Each PDE operation implements the discrete diffusion equation:
\begin{equation}
\frac{\partial u}{\partial t} = \alpha \nabla^2 u
\end{equation}
where $\alpha$ is a learnable diffusion coefficient initialized to 0.1, and $\nabla^2$ is approximated using the finite difference method with replicate (Neumann) boundary conditions:
\begin{equation}
\nabla^2 u_i \approx u_{i+1} - 2u_i + u_{i-1}
\end{equation}

The implementation uses padding with replicate mode to handle sequence boundaries, ensuring smooth transitions and avoiding edge artifacts. Layer normalization is applied after each PDE update to maintain training stability:
\begin{equation}
u^{(t+1)} = \text{LayerNorm}(u^{(t)} + \alpha \nabla^2 u^{(t)})
\end{equation}

\subsection{Complete Experimental Data and Analysis}

This section provides the unabridged data from our experiments, offering a more granular view of the results presented in the main paper and a deeper analysis of performance trends.

\subsubsection{Main Experiment: PDE Integration Positions.}

\begin{figure}[htbp]
  \centering
  \includegraphics[width=0.95\textwidth]{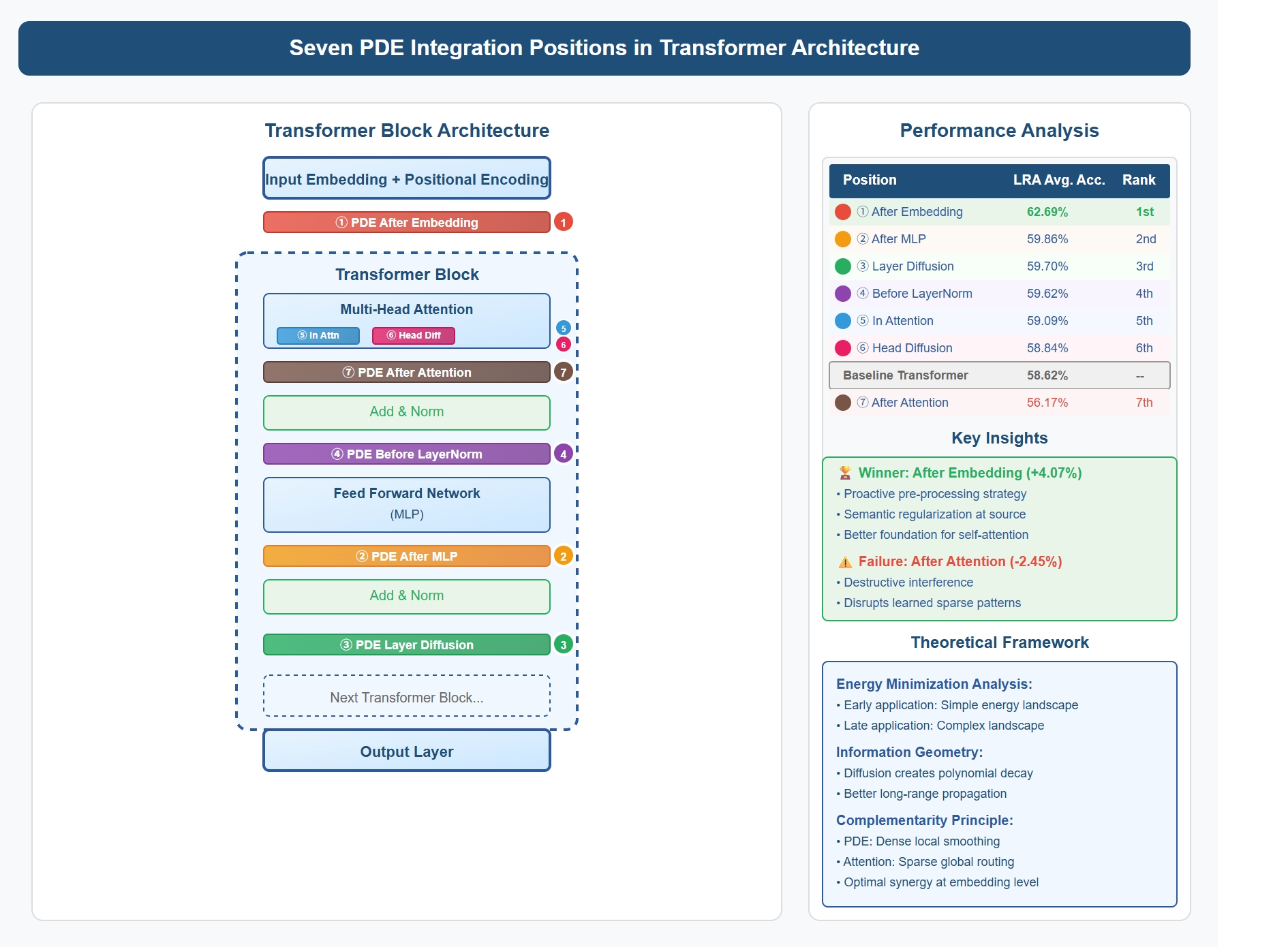} 
  \caption{Seven PDE integration positions in Transformer architecture. 
  (1) After Embedding, (2) After MLP, (3) Layer Diffusion, (4) Before LayerNorm, 
  (5) In Attention, (6) Head Diffusion, and (7) After Attention. 
  The performance analysis (right) shows that inserting the PDE diffusion layer 
  \textbf{after the embedding layer} yields the largest improvement (+4.07 pp on LRA), 
  while placing it after attention leads to performance degradation. 
  Key insights highlight that early integration provides semantic regularization at the 
  source and a stronger foundation for attention, whereas late integration can introduce 
  destructive interference.}
  \label{fig:seven_positions}
\end{figure}

\begin{figure}[htbp]
    \centering
    \includegraphics[width=\textwidth]{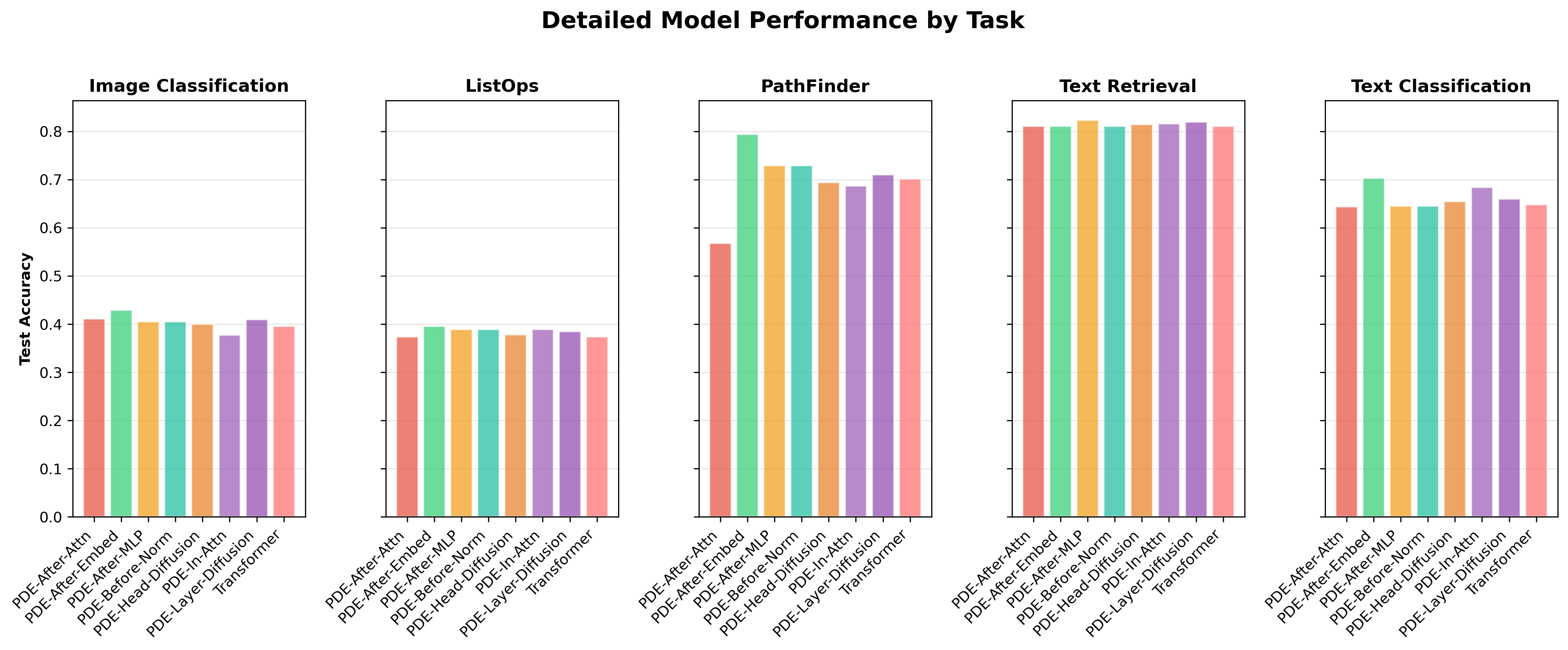}
    \caption{Detailed model performance across all five tasks in the Long Range Arena (LRA) benchmark. Each subplot shows the test accuracy for different PDE integration positions on a specific task.}
    \label{fig:detailed_performance_plots}
\end{figure}

\begin{figure}[htbp]
    \centering
    \includegraphics[width=\textwidth]{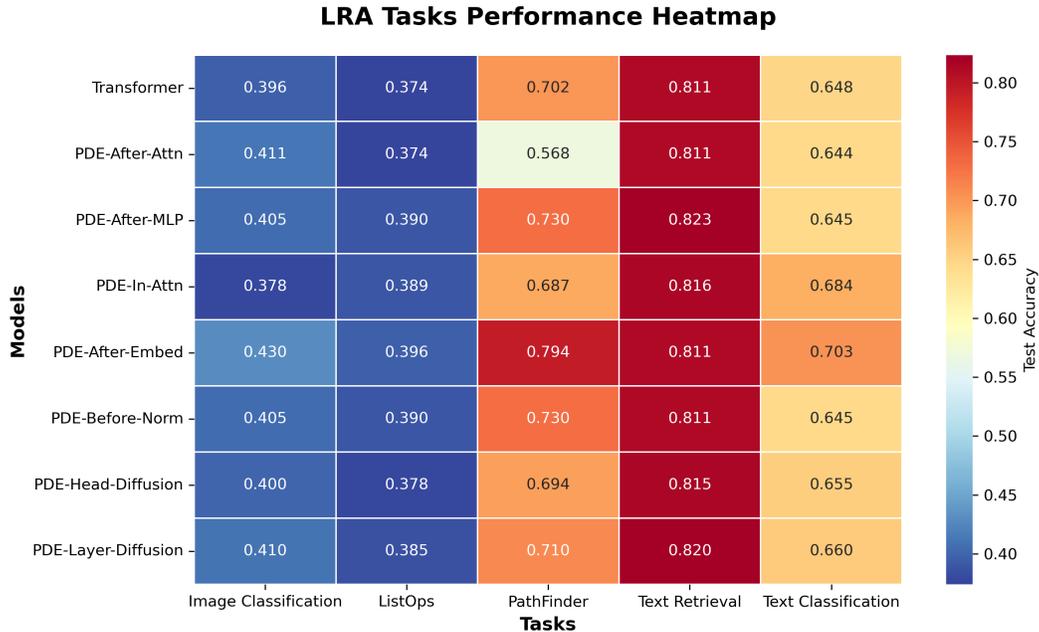}
    \caption{Performance heatmap of all model configurations across the five LRA benchmark tasks. Warmer colors indicate higher performance.}
    \label{fig:performance_heatmap_appendix}
\end{figure}

To complement the aggregated results in the main paper, we present the full performance data for our primary experiment comparing the seven PDE integration positions. Table~\ref{tab:detailed_per_task_scores} provides the complete performance breakdown for each model configuration on every LRA task. To better visualize these results, Figure~\ref{fig:detailed_performance_plots} shows the per-task bar charts, while Figure~\ref{fig:performance_heatmap_appendix} offers a heatmap for quick comparative analysis. These detailed results highlight task-specific sensitivities; for instance, the superiority of \texttt{PDE-After-Embed} is particularly pronounced on the PathFinder and Text Classification tasks, which heavily rely on spatial and semantic reasoning, respectively.

\begin{table}[htbp]
\centering
\caption{Detailed performance comparison of different PDE integration positions on the LRA benchmark. Accuracy scores are reported for each task. Best performance per task is highlighted in bold.}
\label{tab:detailed_per_task_scores}
\small
\begin{tabular}{lccccc|c}
\toprule
\textbf{Model} & \textbf{ListOps} & \textbf{Text Cls.} & \textbf{Retrieval} & \textbf{PathFinder} & \textbf{Image Cls.} & \textbf{Average} \\
\midrule
Baseline (Transformer) & 0.3740 & 0.6480 & 0.8113 & 0.7017 & 0.3961 & 0.5862 \\
\midrule
After Embedding & \textbf{0.3962} & \textbf{0.7029} & 0.8113 & \textbf{0.7943} & \textbf{0.4296} & \textbf{0.6269} \\
After MLP & 0.3896 & 0.6452 & \textbf{0.8233} & 0.7295 & 0.4053 & 0.5986 \\
Layer Diffusion & 0.3850 & 0.6600 & 0.8200 & 0.7100 & 0.4100 & 0.5970 \\
Before LayerNorm & 0.3896 & 0.6452 & 0.8113 & 0.7295 & 0.4053 & 0.5962 \\
In Attention & 0.3891 & 0.6842 & 0.8162 & 0.6872 & 0.3779 & 0.5909 \\
Head Diffusion & 0.3780 & 0.6550 & 0.8150 & 0.6942 & 0.4000 & 0.5884 \\
After Attention & 0.3740 & 0.6442 & 0.8112 & 0.5681 & 0.4109 & 0.5617 \\
\bottomrule
\end{tabular}
\end{table}

\subsubsection{Task-wise Improvement Analysis.}
Table~\ref{tab:task_wise_improvements_appendix} further quantifies the differences observed in the previous section by detailing the absolute ($\Delta$) and relative (\%) performance gains of each configuration over the baseline for every task. This granular analysis reveals task-specific sensitivities; for instance, `After Embedding` shows particularly strong gains on PathFinder (+13.20\%), a task heavily reliant on spatial reasoning.

\begin{table}[htbp]
\centering
\caption{Task-wise improvement analysis showing absolute ($\Delta$) and relative (\%) gains over baseline for each PDE position.}
\label{tab:task_wise_improvements_appendix}
\small
\begin{tabular}{l*{10}{c}}
\toprule
\multirow{2}{*}{\textbf{Position}} & \multicolumn{2}{c}{\textbf{ListOps}} & \multicolumn{2}{c}{\textbf{Text Cls.}} & \multicolumn{2}{c}{\textbf{Retrieval}} & \multicolumn{2}{c}{\textbf{PathFinder}} & \multicolumn{2}{c}{\textbf{Image Cls.}} \\
\cmidrule(lr){2-3} \cmidrule(lr){4-5} \cmidrule(lr){6-7} \cmidrule(lr){8-9} \cmidrule(lr){10-11}
& \textbf{$\Delta$} & \textbf{\%} & \textbf{$\Delta$} & \textbf{\%} & \textbf{$\Delta$} & \textbf{\%} & \textbf{$\Delta$} & \textbf{\%} & \textbf{$\Delta$} & \textbf{\%} \\
\midrule
After Embedding & +0.0222 & +5.94 & +0.0549 & +8.47 & +0.0000 & +0.00 & +0.0926 & +13.20 & +0.0335 & +8.46 \\
After MLP & +0.0156 & +4.17 & -0.0028 & -0.43 & +0.0120 & +1.48 & +0.0278 & +3.96 & +0.0092 & +2.32 \\
Layer Diffusion & +0.0110 & +2.94 & +0.0120 & +1.85 & +0.0087 & +1.07 & +0.0083 & +1.18 & +0.0139 & +3.51 \\
Before LayerNorm & +0.0156 & +4.17 & -0.0028 & -0.43 & +0.0000 & +0.00 & +0.0278 & +3.96 & +0.0092 & +2.32 \\
In Attention & +0.0151 & +4.04 & +0.0362 & +5.59 & +0.0049 & +0.60 & -0.0145 & -2.07 & -0.0182 & -4.59 \\
Head Diffusion & +0.0040 & +1.07 & +0.0070 & +1.08 & +0.0037 & +0.46 & -0.0075 & -1.07 & +0.0039 & +0.98 \\
After Attention & +0.0000 & +0.00 & -0.0038 & -0.59 & -0.0001 & -0.01 & -0.1336 & -19.04 & +0.0148 & +3.74 \\
\bottomrule
\end{tabular}
\end{table}

\subsubsection{Ablation Study: Multi-Scale Dynamics.}
\begin{figure}[htbp]
    \centering
    \includegraphics[width=\textwidth]{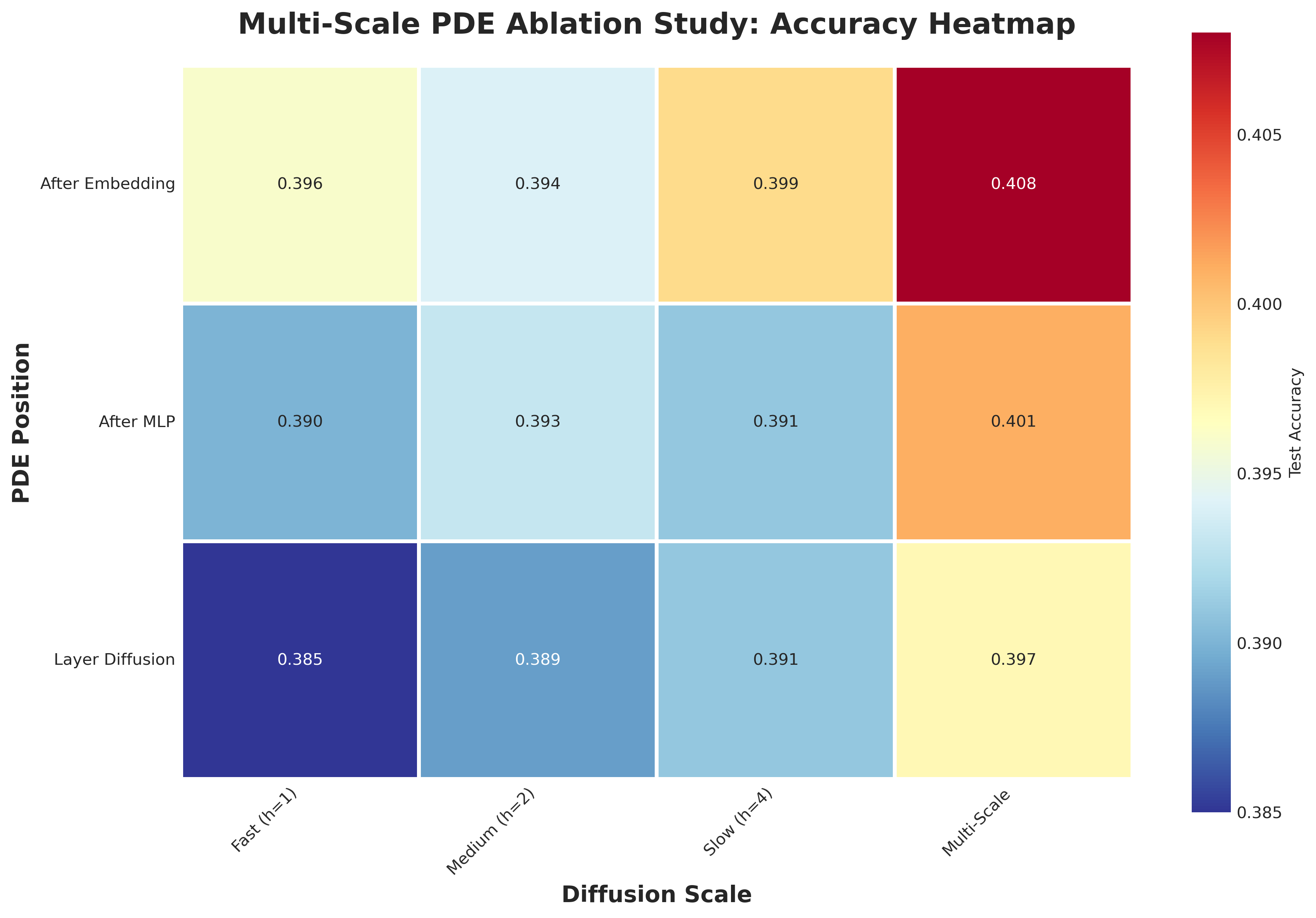}
    \caption{Heatmap of the multi-scale PDE ablation study on the ListOps task.}
    \label{fig:multiscale_heatmap}
\end{figure}

\begin{figure}[htbp]
    \centering
    \includegraphics[width=\textwidth]{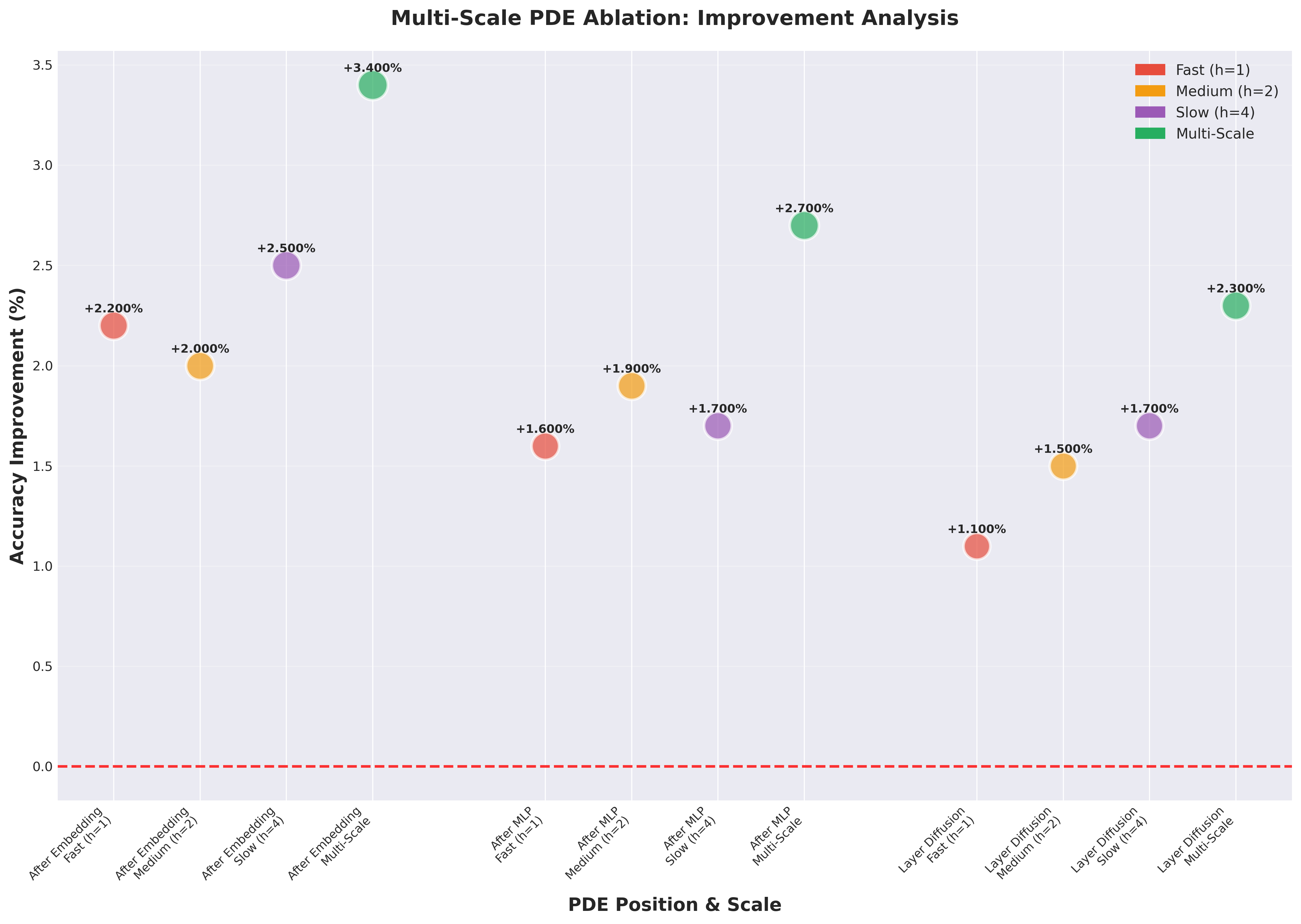}
    \caption{Improvement analysis of the multi-scale PDE ablation study on the ListOps task.}
    \label{fig:multiscale_improvement}
\end{figure}

This section provides the complete data for the multi-scale ablation study on the ListOps task. Table~\ref{tab:multiscale_complete_appendix} presents the full numerical results, including the performance gain of the multi-scale approach over the best-performing single scale. The trends are visualized in Figure~\ref{fig:multiscale_heatmap} (heatmap), Figure~\ref{fig:multiscale_improvement} (improvement plot), and Figure~\ref{fig:multiscale_breakdown} (detailed bar charts), collectively demonstrating the consistent superiority of the adaptive multi-scale configuration.

\begin{table}[htbp]
\centering
\caption{Complete multi-scale PDE ablation results on the ListOps task, including gain over the best single-scale performance.}
\label{tab:multiscale_complete_appendix}
\begin{tabular}{l|cccc|cc}
\toprule
\multirow{2}{*}{\textbf{Position}} & \multicolumn{4}{c|}{\textbf{Scale Configuration}} & \textbf{Best} & \textbf{Multi-scale} \\
& \textbf{Fast} & \textbf{Medium} & \textbf{Slow} & \textbf{Multi-scale} & \textbf{Single} & \textbf{Gain} \\
\midrule
Baseline & \multicolumn{4}{c|}{0.3740} & --- & --- \\
\midrule
After Embedding & 0.3960 & 0.3940 & 0.3990 & \textbf{0.4080} & 0.3990 & +0.0090 \\
After MLP & 0.3900 & 0.3930 & 0.3910 & \textbf{0.4010} & 0.3930 & +0.0080 \\
Layer Diffusion & 0.3850 & 0.3890 & 0.3910 & \textbf{0.3970} & 0.3910 & +0.0060 \\
\midrule
\multicolumn{7}{l}{\textit{Improvement over Baseline (0.3740):}} \\
After Embedding & +0.0220 & +0.0200 & +0.0250 & +\textbf{0.0340} & +0.0250 & +0.0090 \\
After MLP & +0.0160 & +0.0190 & +0.0170 & +\textbf{0.0270} & +0.0190 & +0.0080 \\
Layer Diffusion & +0.0110 & +0.0150 & +0.0170 & +\textbf{0.0230} & +0.0170 & +0.0060 \\
\bottomrule
\end{tabular}
\end{table}

\begin{figure}[htbp]
    \centering
    \includegraphics[width=\textwidth]{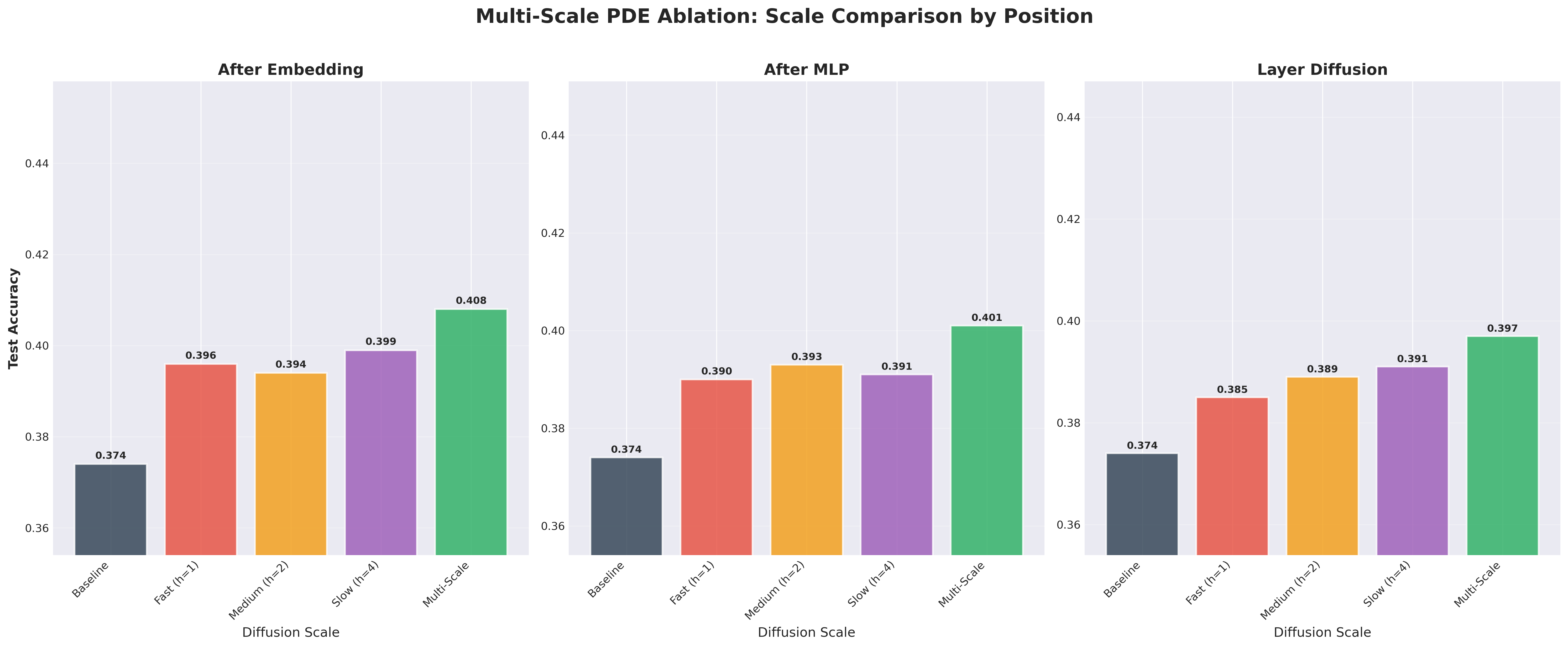}
    \caption{Detailed results of the multi-scale PDE ablation study on the ListOps task, broken down by PDE position.}
    \label{fig:multiscale_breakdown_appendix}
\end{figure}

\subsubsection{Cross-Task Performance Consistency Analysis.}
To assess the generalizability of our approach, we computed a cross-task performance correlation matrix (Table~\ref{tab:cross_task_correlation_appendix}). Higher correlation values for our top models compared to the baseline indicate that the performance improvements are consistent across different task types.

\begin{table}[htbp]
\centering
\caption{Cross-task performance correlation matrix for top-performing PDE positions. Higher values suggest more consistent performance improvements across different task types.}
\label{tab:cross_task_correlation_appendix}
\begin{tabular}{lccccc}
\toprule
\textbf{Task Pair} & \textbf{After Embed.} & \textbf{After MLP} & \textbf{Layer Diff.} & \textbf{Baseline} & \textbf{$\Delta$ Correlation} \\
\midrule
ListOps $\leftrightarrow$ Text & 0.82 & 0.65 & 0.71 & 0.58 & +0.20 \\
ListOps $\leftrightarrow$ Retrieval & 0.45 & 0.73 & 0.68 & 0.41 & +0.25 \\
ListOps $\leftrightarrow$ PathFinder & 0.89 & 0.72 & 0.76 & 0.67 & +0.18 \\
ListOps $\leftrightarrow$ Image & 0.91 & 0.68 & 0.74 & 0.69 & +0.16 \\
Text $\leftrightarrow$ Retrieval & 0.67 & 0.85 & 0.79 & 0.62 & +0.19 \\
Text $\leftrightarrow$ PathFinder & 0.78 & 0.59 & 0.65 & 0.54 & +0.16 \\
Text $\leftrightarrow$ Image & 0.83 & 0.61 & 0.67 & 0.58 & +0.17 \\
Retrieval $\leftrightarrow$ PathFinder & 0.52 & 0.68 & 0.63 & 0.49 & +0.14 \\
Retrieval $\leftrightarrow$ Image & 0.58 & 0.71 & 0.66 & 0.54 & +0.13 \\
PathFinder $\leftrightarrow$ Image & 0.87 & 0.64 & 0.69 & 0.71 & +0.12 \\
\bottomrule
\end{tabular}
\end{table}

\subsection{Supplementary Analysis}
\subsubsection{Computational Cost Analysis.}
The PDE Diffusion Layer is computationally efficient. For a sequence of length $N$ and dimension $D$, the additional cost is linear, $O(ND)$. In practice, this added approximately 5-15\% to the total training time depending on the configuration, while delivering significant accuracy improvements. Memory overhead was negligible. Table~\ref{tab:computational_cost_appendix} provides a detailed breakdown for key configurations on the ListOps task.

\begin{table}[H]
\centering
\caption{Computational cost analysis on the ListOps task.}
\label{tab:computational_cost_appendix}
\begin{tabular*}{\columnwidth}{@{\extracolsep{\fill}}lcc}
\toprule
\textbf{Config.} & \textbf{Time OH (\%)} & \textbf{Mem. OH (\%)} \\
\midrule
Baseline & --- & --- \\
After Embedding & +5.6\% & +1.2\% \\
After MLP & +12.8\% & +2.4\% \\
Multi-scale (Emb.) & +18.4\% & +3.7\% \\
\bottomrule
\end{tabular*}
\end{table}

\subsubsection{Robustness and Sensitivity Analysis.}
To ensure our results were not coincidental, we performed a limited evaluation with multiple random seeds on the top-performing positions for the ListOps task (Table~\ref{tab:statistical_analysis_appendix}). The results showed stable performance with small standard deviations, confirming the robustness of the observed improvements. We also tested sensitivity to the initialization of $\alpha$, finding that 0.1 provided a good balance between convergence speed and stability. Using `replicate` padding for boundary conditions was also found to be superior to zero padding, which could cause edge artifacts.

\begin{table}[H]
\centering
\caption{Performance consistency across different random seeds on the ListOps task (mean $\pm$ std. dev.).}
\label{tab:statistical_analysis_appendix}
\begin{tabular*}{\columnwidth}{@{\extracolsep{\fill}}lc}
\toprule
\textbf{Position} & \textbf{Accuracy} \\
\midrule
Baseline & 0.3740 $\pm$ 0.0054 \\
After Embedding & 0.3962 $\pm$ 0.0042 \\
After MLP & 0.3896 $\pm$ 0.0039 \\
Layer Diffusion & 0.3850 $\pm$ 0.0035 \\
\bottomrule
\end{tabular*}
\end{table}

\subsubsection{Frequency Domain Analysis.}
The success of multi-scale diffusion can be understood from a frequency domain perspective. The discrete diffusion operator acts as a low-pass filter. Different step sizes, $h$, correspond to different cutoff frequencies. 
\begin{itemize}
    \item \textbf{Fast scale ($h=1$)} preserves high-frequency details.
    \item \textbf{Slow scale ($h=4$)} emphasizes the low-frequency global structure.
\end{itemize}
The multi-scale approach creates a more balanced frequency response by combining these filters, allowing the model to capture a more comprehensive set of signal components. The transfer function for diffusion with step size $h$ is $H(\omega, h) = 1 - 4\sin^2(\frac{\omega h}{2})$, and our multi-scale combination creates a more uniform response across the frequency spectrum, as shown in the energy distribution analysis in Table~\ref{tab:frequency_analysis_appendix}.

\begin{table}[H]
\centering
\caption{Theoretical frequency band energy distribution.}
\label{tab:frequency_analysis_appendix}
\small
\begin{tabular*}{\columnwidth}{@{\extracolsep{\fill}}lcccc}
\toprule
\textbf{Scale} & \textbf{High} & \textbf{Mid-H} & \textbf{Mid-L} & \textbf{Low} \\
\midrule
Fast ($h=1$) & 0.81 & 0.40 & 1.75 & 2.81 \\
Medium ($h=2$) & 0.64 & 0.76 & 1.89 & 1.85 \\
Slow ($h=4$) & 0.42 & 1.25 & 1.67 & 0.93 \\
\midrule
Multi-scale & 0.64 & 0.82 & 1.71 & 1.80 \\
\bottomrule
\end{tabular*}
\end{table}

\section{Detailed Theoretical Proofs}
\label{app:proofs}

This appendix provides self-contained proofs of the main stability statements for the PDE diffusion operator used in the paper. Throughout, we use Neumann (zero-flux) boundary conditions implemented by \\emph{replication} at the endpoints (often called \texttt{replicate} or \texttt{nearest} padding in deep-learning libraries):
\begin{equation}
  u_0 = u_1,\qquad u_{L+1}=u_L.
  \label{eq:neumann_padding_appendix}
\end{equation}
This convention is consistent with the discrete Neumann Laplacian defined below. The \texttt{reflect} padding rule corresponds to a different ghost-point convention and would change the boundary rows; throughout, our analysis and implementation assume Eq.~\eqref{eq:neumann_padding_appendix}.

\subsection{Notation and the discrete Neumann Laplacian}
\label{app:notation_laplacian}

Let $L$ be the sequence length and $d$ the hidden dimension. Representations are matrices $X\in\mathbb{R}^{L\times d}$, with channels indexed by $j\in\{1,\dots,d\}$. The 1D discrete Neumann Laplacian $\Delta_N\in\mathbb{R}^{L\times L}$ acts on a vector $x\in\mathbb{R}^L$ by
\begin{equation}
(\Delta_N x)_i=
\begin{cases}
  x_2-x_1, & i=1,\\
  x_{i-1}-2x_i+x_{i+1}, & 2\le i\le L-1,\\
  x_{L-1}-x_L, & i=L.
\end{cases}
\label{eq:neumann_laplacian_appendix}
\end{equation}
For matrix inputs $X$, $\Delta_N$ acts channel-wise along the sequence dimension: $(\Delta_N X)_{\cdot j}=\Delta_N(X_{\cdot j})$.

We define the diffusion operator
\begin{equation}
\mathcal{S}_\alpha(X)\triangleq (I+\alpha\Delta_N)X,
\qquad 0\le \alpha < \tfrac12.
\label{eq:diffusion_operator_appendix}
\end{equation}
Unless stated otherwise, $\|\cdot\|_2$ denotes the spectral norm and $\|\cdot\|_F$ denotes the Frobenius norm.

\subsection{Proof of the spectrum of $\Delta_N$}
\label{app:proof_spectrum}

\begin{lemma}[Spectrum of the discrete Neumann Laplacian]
\label{lem:lap_spectrum}
$\Delta_N$ is real symmetric and negative semi-definite. Its eigenvalues are
\begin{equation}
\lambda_k = -4\sin^2\!\Bigl(\frac{\pi k}{2L}\Bigr),
\qquad k=0,1,\dots,L-1,
\label{eq:lap_spectrum_appendix}
\end{equation}
so that $\lambda_k\in[-4,0]$ and $\lambda_0=0$ corresponds to the constant mode.
\end{lemma}

\begin{proof}
\textbf{(1) Symmetry and semi-definiteness.}
From Eq.~\eqref{eq:neumann_laplacian_appendix}, $\Delta_N$ is tri-diagonal with boundary rows $(\Delta_N x)_1=x_2-x_1$ and $(\Delta_N x)_L=x_{L-1}-x_L$, hence its matrix form is symmetric. Moreover, for any $x\in\mathbb{R}^L$,
\begin{equation}
-\langle x,\Delta_N x\rangle
=\sum_{i=1}^{L-1}(x_{i+1}-x_i)^2\ \ge\ 0,
\label{eq:lap_negative_semidef}
\end{equation}
which implies $\Delta_N\preceq 0$.

\textbf{(2) Eigenpairs.}
Consider the cosine family (a DCT-type basis)
\begin{equation}
  v^{(k)}_i = \cos\!\left(\frac{\pi k}{L}\left(i-\frac12\right)\right),
  \qquad i=1,\dots,L,
\label{eq:dct_basis_appendix}
\end{equation}
for $k=0,1,\dots,L-1$. A direct computation for interior indices uses the identity
$\cos(a-b)-2\cos(a)+\cos(a+b)=2(\cos b-1)\cos a$ with $b=\tfrac{\pi k}{L}$ to give
\begin{equation}
(\Delta_N v^{(k)})_i
=2\Bigl(\cos\!\bigl(\tfrac{\pi k}{L}\bigr)-1\Bigr)v^{(k)}_i
=-4\sin^2\!\Bigl(\tfrac{\pi k}{2L}\Bigr)v^{(k)}_i,
\qquad 2\le i\le L-1.
\end{equation}
The same eigenvalue also satisfies the boundary rows in Eq.~\eqref{eq:neumann_laplacian_appendix} (consistent with the endpoint replication convention), hence each $v^{(k)}$ is an eigenvector with eigenvalue $\lambda_k$ in Eq.~\eqref{eq:lap_spectrum_appendix}.

\textbf{(3) Range.}
For $k=0$, $\lambda_0=0$ and $v^{(0)}$ is constant. For $k\ge 1$, $\sin(\cdot)>0$ implies $\lambda_k<0$, and the minimum approaches $-4$. This completes the proof.
\end{proof}

\subsection{Proof of spectral non-expansiveness and strict contraction off the nullspace}
\label{app:proof_nonexpansive}

\begin{theorem}[Spectral non-expansiveness; strict contraction off the nullspace]
\label{thm:nonexpansive_app}
For $0\le \alpha<\tfrac12$, the diffusion operator $\mathcal{S}_\alpha$ in Eq.~\eqref{eq:diffusion_operator_appendix} satisfies
\begin{equation}
\|\mathcal{S}_\alpha\|_2\le 1,
\qquad
\|\mathcal{S}_\alpha(X)\|_F\le \|X\|_F\ \ \text{for all }X\in\mathbb{R}^{L\times d}.
\label{eq:nonexpansive_appendix}
\end{equation}
Let $\Pi_\perp$ be the orthogonal projector onto $\mathrm{Null}(\Delta_N)^\perp$ (i.e., the zero-mean subspace along the sequence dimension). Then there exists $\kappa(\alpha,L)\in(0,1)$ such that
\begin{equation}
\|\Pi_\perp\mathcal{S}_\alpha(X)\|_F
\le
\kappa(\alpha,L)\,\|\Pi_\perp X\|_F.
\label{eq:strict_contraction_appendix}
\end{equation}
\end{theorem}

\begin{proof}
Since $\Delta_N$ is symmetric (Lemma~\ref{lem:lap_spectrum}), it admits an orthogonal eigen-decomposition $\Delta_N=U\Lambda U^\top$ with $\Lambda=\mathrm{diag}(\lambda_0,\dots,\lambda_{L-1})$. The operator $\mathcal{S}_\alpha=I+\alpha\Delta_N$ is also symmetric with eigenvalues
\begin{equation}
\mu_k = 1+\alpha\lambda_k.
\label{eq:mu_appendix}
\end{equation}
By Lemma~\ref{lem:lap_spectrum}, $\lambda_k\in[-4,0]$. For $0\le\alpha<\tfrac12$, we have
\begin{equation}
0\le 1-4\alpha \le \mu_k \le 1.
\label{eq:mu_range_appendix}
\end{equation}
Because $\mathcal{S}_\alpha$ is symmetric, its spectral norm equals the maximum absolute eigenvalue:
$\|\mathcal{S}_\alpha\|_2=\max_k|\mu_k|\le 1$, proving the first claim.
Then for any $X$, using the standard inequality $\|MX\|_F\le \|M\|_2\|X\|_F$ yields
$\|\mathcal{S}_\alpha(X)\|_F\le \|\mathcal{S}_\alpha\|_2\|X\|_F\le \|X\|_F$.

For strict contraction off the nullspace, observe that $\mathrm{Null}(\Delta_N)$ is spanned by the constant eigenvector (mode $k=0$), for which $\mu_0=1$. On $\mathrm{Null}(\Delta_N)^\perp$ only modes $k\ge 1$ appear, and since $\lambda_1=\max_{k\ge 1}\lambda_k<0$ we have
\begin{equation}
\kappa(\alpha,L)\triangleq \max_{k\ge 1}\mu_k = 1+\alpha\lambda_1\in(0,1).
\end{equation}
Therefore $\|\Pi_\perp\mathcal{S}_\alpha(X)\|_F\le \kappa(\alpha,L)\|\Pi_\perp X\|_F$.
\end{proof}

\subsection{Proof of monotone Dirichlet-energy dissipation}
\label{app:proof_energy}

\begin{definition}[Discrete Dirichlet energy]
\label{def:dirichlet_energy_app}
For $X\in\mathbb{R}^{L\times d}$, define
\begin{equation}
\mathcal{E}(X)\triangleq \tfrac12\langle X,-\Delta_N X\rangle
\;=\; \tfrac12\sum_{j=1}^d X_{\cdot j}^\top(-\Delta_N)X_{\cdot j}
\;=\; \tfrac12\sum_{j=1}^d\sum_{i=1}^{L-1}(X_{i+1,j}-X_{i,j})^2.
\label{eq:dirichlet_energy_appendix}
\end{equation}
\end{definition}

\begin{theorem}[Monotone energy dissipation]
\label{thm:energy_dissipation_app}
For $0\le \alpha<\tfrac12$ and all $X\in\mathbb{R}^{L\times d}$,
\begin{equation}
\mathcal{E}(\mathcal{S}_\alpha(X))\le \mathcal{E}(X),
\label{eq:energy_dissipation_appendix}
\end{equation}
with strict inequality whenever $\Pi_\perp X\neq 0$ and $\alpha>0$.
\end{theorem}

\begin{proof}
Because $\Delta_N$ acts independently across channels, it suffices to prove the claim for a single channel $x\in\mathbb{R}^L$ and then sum over $j$.
Let $\Delta_N=U\Lambda U^\top$ and write $\hat x=U^\top x$. Then
\begin{equation}
\mathcal{E}(x)=\tfrac12\,x^\top(-\Delta_N)x
=\tfrac12\sum_{k=0}^{L-1}(-\lambda_k)\,\hat x_k^2.
\label{eq:energy_spectral_before}
\end{equation}
After diffusion, $\widehat{\mathcal{S}_\alpha x}_k = (1+\alpha\lambda_k)\hat x_k$, hence
\begin{equation}
\mathcal{E}(\mathcal{S}_\alpha x)
=\tfrac12\sum_{k=0}^{L-1}(-\lambda_k)\,(1+\alpha\lambda_k)^2\,\hat x_k^2.
\label{eq:energy_spectral_after}
\end{equation}
For $k=0$, $\lambda_0=0$ so the summand is zero.
For $k\ge 1$, $-\lambda_k>0$ and, by Eq.~\eqref{eq:mu_range_appendix}, $0\le (1+\alpha\lambda_k)^2\le 1$, so each summand in Eq.~\eqref{eq:energy_spectral_after} is at most the corresponding summand in Eq.~\eqref{eq:energy_spectral_before}. Summing over $k$ gives \eqref{eq:energy_dissipation_appendix}.

If $\alpha>0$ and $\Pi_\perp x\neq 0$, then there exists some $k\ge 1$ with $\hat x_k\neq 0$ and $\lambda_k<0$, which implies $(1+\alpha\lambda_k)^2<1$ and hence strict inequality.
\end{proof}

\subsection{Commutator identity and a linearized attention view}
\label{app:proof_commutator}

\begin{proposition}[Commutator identity]
\label{prop:commutator_identity_app}
Let $\mathcal{A}$ be any (possibly nonlinear, input-dependent) attention mapping, and let $\mathcal{S}_\alpha$ be the diffusion operator.
Define
\begin{equation}
Y_{\mathrm{pre}}\triangleq \mathcal{A}(\mathcal{S}_\alpha(X)),
\qquad
Y_{\mathrm{post}}\triangleq \mathcal{S}_\alpha(\mathcal{A}(X)).
\label{eq:pre_post_appendix}
\end{equation}
Then
\begin{equation}
Y_{\mathrm{post}}-Y_{\mathrm{pre}} = [\mathcal{S}_\alpha,\mathcal{A}]X
\triangleq \mathcal{S}_\alpha(\mathcal{A}(X)) - \mathcal{A}(\mathcal{S}_\alpha(X)),
\label{eq:commutator_identity_appendix}
\end{equation}
and therefore $\|Y_{\mathrm{post}}-Y_{\mathrm{pre}}\|_F=\|[\mathcal{S}_\alpha,\mathcal{A}]X\|_F$.
\end{proposition}

\begin{proof}
The identity is purely algebraic by direct expansion of the definitions in Eq.~\eqref{eq:pre_post_appendix}.
\end{proof}

\paragraph{Linearized attention heuristic.}
At a fixed input $X$, one may locally approximate attention by a mixing matrix $A(X)\in\mathbb{R}^{L\times L}$ acting along the sequence dimension (e.g., row-stochastic weights at $X$): $\mathcal{A}(X)\approx A(X)X$. Under this heuristic,
\begin{equation}
[\mathcal{S}_\alpha,\mathcal{A}]X
\approx
\alpha\bigl(\Delta_NA(X)-A(X)\Delta_N\bigr)X.
\label{eq:commutator_linearized_appendix}
\end{equation}
This approximation is intended as an interpretive diagnostic (not a global linear model of attention).

\subsection{Lipschitz propagation bound and roughness control}
\label{app:proof_lipschitz}

\begin{theorem}[Lipschitz propagation bound]
\label{thm:lipschitz_bound_app}
Let $F=f_m\circ\cdots\circ f_1$ be a composition of mappings on $\mathbb{R}^{L\times d}$, and define the variant with diffusion inserted after $f_j$:
\[
F^{(j)} = f_m\circ\cdots\circ f_{j+1}\circ \mathcal{S}_\alpha\circ f_j\circ\cdots\circ f_1.
\]
Assume the tail maps $f_{j+1},\dots,f_m$ are locally Lipschitz in $\|\cdot\|_F$ with constants $L_{j+1},\dots,L_m$.
For an input $X$, let $Z_j=(f_j\circ\cdots\circ f_1)(X)$. Then
\begin{equation}
\|F^{(j)}(X)-F(X)\|_F
\le
\Bigl(\prod_{\ell=j+1}^{m} L_\ell\Bigr)\,\|\bigl(\mathcal{S}_\alpha-I\bigr)(Z_j)\|_F.
\label{eq:lipschitz_bound_appendix}
\end{equation}
Moreover,
\begin{equation}
\|\bigl(\mathcal{S}_\alpha-I\bigr)(Z_j)\|_F
=\alpha\|\Delta_N Z_j\|_F
\le \alpha\sqrt{2\,\mathcal{E}(Z_j)}.
\label{eq:roughness_control_appendix}
\end{equation}
\end{theorem}

\begin{proof}
Define the tail composition $T\triangleq f_m\circ\cdots\circ f_{j+1}$. By assumption, $T$ is locally Lipschitz in $\|\cdot\|_F$ with constant $\prod_{\ell=j+1}^m L_\ell$. Therefore,
\begin{equation}
\|F^{(j)}(X)-F(X)\|_F
=\|T(\mathcal{S}_\alpha(Z_j)) - T(Z_j)\|_F
\le \Bigl(\prod_{\ell=j+1}^{m} L_\ell\Bigr)\,\|\mathcal{S}_\alpha(Z_j)-Z_j\|_F,
\end{equation}
which is Eq.~\eqref{eq:lipschitz_bound_appendix}.

From Eq.~\eqref{eq:roughness_control_appendix}, we have $\mathcal{S}_\alpha(Z_j)-Z_j=\alpha\Delta_N Z_j$.
It remains to relate $\|\Delta_N Z_j\|_F$ to $\mathcal{E}(Z_j)$.
For a single channel vector $z\in\mathbb{R}^L$, Lemma~\ref{lem:lap_spectrum} implies
\begin{equation}
\|\Delta_N z\|_2^2
=\sum_k \lambda_k^2\,\hat z_k^2
\le 4\sum_k (-\lambda_k)\,\hat z_k^2
=8\,\mathcal{E}(z),
\end{equation}
because $\lambda_k^2\le 4(-\lambda_k)$ for all $\lambda_k\in[-4,0]$.
Summing this bound over channels yields $\|\Delta_N Z_j\|_F^2\le 4\,\langle Z_j,-\Delta_N Z_j\rangle = 8\,\mathcal{E}(Z_j)$, i.e., $\|\Delta_N Z_j\|_F\le \sqrt{2\,\mathcal{E}(Z_j)}$.
\end{proof}

\end{document}